%% file: tmi.tex
\documentclass[journal,twoside,web]{ieeecolor}
\usepackage{tmi}
\usepackage{cite}
\usepackage{amsmath,amssymb,amsfonts}
\usepackage{algorithmic}
\usepackage{graphicx}
\usepackage{textcomp}
\usepackage{url}
\usepackage{xspace}
\usepackage{multirow}

\usepackage{amsmath}
\usepackage{subcaption}
\usepackage{graphicx}
\usepackage{colortbl}
\usepackage{caption}
\usepackage{tabularx}
\usepackage{wrapfig}
\usepackage{makecell}
\usepackage{xcolor}
\usepackage{ulem}
\usepackage{hyperref}       % hyperlinks
\usepackage{cleveref}
\usepackage{url}            % simple URL typesetting
\usepackage{booktabs}       % professional-quality tables
\usepackage{amsfonts}       % blackboard math symbols
\usepackage{nicefrac}       % compact symbols for 1/2, etc.
\usepackage{microtype}      % microtypography
\usepackage{algorithm}
\usepackage{algorithmic}

\newcommand{\jc}[1]{\textcolor{black}{#1}}

\newcommand{\cmx}[1]{\textcolor{black}{#1}}
\definecolor{revcolor}{RGB}{0,110,105}
\definecolor{revcolor}{RGB}{0,0,0}
\long\def\rev#1{{\color{revcolor}#1}}

\def\BibTeX{{\rm B\kern-.05em{\sc i\kern-.025em b}\kern-.08em
    T\kern-.1667em\lower.7ex\hbox{E}\kern-.125emX}}
\begin{document}
\title{Dementia Etiology Diagnosis via Collaborative Meta Knowledge Enhancement}
\author{Siyuan Du, Mengxi Chen, Xinyang Jiang, Zilong Wang, Jiangchao Yao, Dongsheng Li, Ya Zhang, Lili Qiu, \IEEEmembership{Fellow, IEEE}, Yanfeng Wang
\thanks{S. Du is with the School of Computer Science, Fudan University.}
\thanks{M. Chen and J. Yao are with the Cooperative Medianet Innovation Center, Shanghai Jiao Tong University. J. Yao was visiting Microsoft Research Asia, Shanghai, China, during the completion of this work.}
\thanks{X. Jiang, Z. Wang, D. Li, and L. Qiu are with Microsoft Research Asia, Shanghai, China.}
\thanks{Ya Zhang and Y. Wang are with the School of Artificial Intelligence, Shanghai Jiao Tong University.}
\thanks{Corresponding authors: J. Yao (sunarker@sjtu.edu.cn) and D. Li (dongsli@microsoft.com).}}

\maketitle

\begin{abstract}
% Artificial intelligence (AI) has shown promising performance in clinical diagnosis, even in some challenging tasks like dementia etiology identification. 
%
\jc{Although artificial intelligence (AI) has shown promising performance in several medical tasks, accurate dementia etiology diagnosis with AI remains challenging due to complex overlapping symptoms among diseases.}
\jc{Scaling up the dataset size by combining the cross-center samples may bring a gain in the pursuit of performance, while the inherent data heterogeneity across centers or populations induces the conflict.}
% Existing models have shown strong performance in AI-assisted dementia diagnosis but struggle with cross-center generalization due to data heterogeneity.
%
Conventional multi-task learning paradigms offer a promising framework; however, they fail to consider critical meta information (\textit{e.g.,} site-specific acquisition and modality availability) to combat the heterogeneity. 
\jc{To address this challenge, we propose a Collaborative Meta Knowledge Enhancement (COME) framework for dementia etiology diagnosis, which} injects multi-center acquisition semantics, source identifiers, and modality indicators as heterogeneity-aware embeddings into a unified Transformer architecture \jc{for scale-up training}, enabling explicit modeling of heterogeneity.
\jc{Besides, a trust-region constrained optimization scheme is designed to regularize the model from spurious correlations during training through a reference model.} 
%
% Across seven independent cohorts, our method achieves state-of-the-art in-domain performance and superior out-of-domain generalization under both cross-center and cross-sequence evaluations.
Across seven independent cohorts, our method achieves state-of-the-art in-domain performance \rev{with a mean macro-averaged AUC of 85.62\% and a 4.29-point gain over the strongest baseline, while maintaining} superior out-of-domain generalization under both cross-center and cross-sequence evaluations.
\jc{Extensive} validation also confirms the alignment between model predictions and established biomarkers (amyloid, tau) and clinical severity, highlighting the potential of COME to enable robust and interpretable dementia diagnostics in real-world settings.

% However, accurate differential AI diagnosis of dementia in multi-center studies is still hindered by substantial imaging heterogeneity, missing modalities, and distribution shifts. 
%

%

\end{abstract}


\input{body/intro}
\input{body/related_works}

\input{body/method}

\input{body/exp}

\input{body/conclusion}

\input{body/appendix}

\input{body/ref}

\end{document}

%% file: body/intro.tex
\section{Introduction}
%整体的逻辑：Explicit problem-solution progression (dementia burden → diagnostic challenge → AI’s promise → multimodal shift → unresolved multi-center gap)

%第一段讲一下现有的挑战，数据异质性，dementia诊断的困难性与挑战性
\jc{Nowadays, dementia is becoming one of the most pressing global health challenges, as nearly 10~million new cases are diagnosed every year~\cite{world2021global}. Accurate and timely identification of the underlying etiology in dementia is crucial for guiding patient management and therapy. However, differential diagnosis of dementia remains challenging due to the substantial overlap in clinical symptoms across etiologies and heterogeneous neuroimaging patterns across individuals and disease stages~\cite{beach2012accuracy}. The complicated disease factors and imaging variability lead to high misdiagnosis rates and ineffective treatment~\cite{hampel2023amyloid}.}

In the past decades, artificial intelligence (AI) has emerged as a promising tool to augment clinical diagnosis~\cite{nawaz2020deep,qiu2022multimodal, xue2024ai,gogishvili2023discovery, li2024lorkd}. Early approaches leveraged unimodal data, such as structural MRI to distinguish \rev{Alzheimer's disease} (AD) from normal aging or mild cognitive impairment (MCI)~\cite{nawaz2020deep}, or cerebrospinal fluid (CSF) biomarkers to predict cognitive decline~\cite{gogishvili2023discovery}. 
However, these single-modality approaches offer an incomplete view of dementia’s multifactorial pathology.  
Recent advances~\cite{qiu2022multimodal, xue2024ai} thus prioritize multimodal integration, combining neuroimaging, clinical assessments, demographics, and biomarkers to enhance diagnostic accuracy. 
\jc{Despite promise, they still fail to achieve satisfactory performance and are limited in the real-world generalization.
A natural strategy, inspired by the recent success in large models, is the scaling law~\cite{jia2021scaling}, which accumulates large-scale data from different sources for training. 
%
% Nevertheless, the heterogeneity across centers or populations induces non-negligible conflicts during training. 
Nevertheless, \rev{when heterogeneous cohorts are simply pooled, differences in acquisition protocols, cohort distributions, and modality coverage can provide inconsistent optimization signals to shared model parameters.}
\rev{The model may thus favor site- or modality-specific correlations that reduce training loss but do not reflect stable disease-related patterns.}
While conventional Multi-Task Learning methods (MTL)~\cite{caruana1993multitask, duong2015low,shazeer2017outrageously,long2017learning,ma2018modeling,senushkin2023independent} can be a potential solution, 
they typically rely on implicit task relationships and lack explicit mechanisms to represent site-specific acquisition protocols or modality availability to combat the heterogeneity. 
This limitation motivates us to explore new ways that can explicitly incorporate such meta knowledge into the training.}

\begin{figure}[t!]
\centerline{\includegraphics[width=0.98\linewidth]{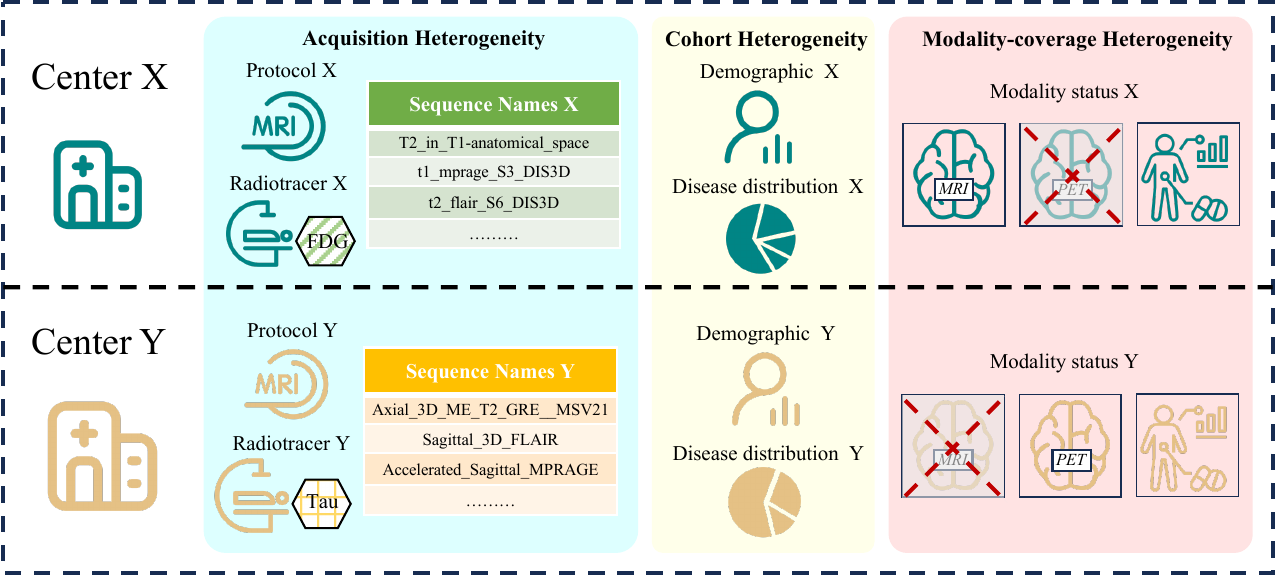}}
\caption {Illustration of heterogeneity across multi-center dementia datasets, exemplified by two sites (Center X and Y).}
\label{fig:intro}
\vspace{-20pt}
\end{figure}

% Large-scale collaborations are essential to capture diverse dementia presentations, but a
As shown in Fig.~\ref{fig:intro}, pooling data across institutions amplifies three compounding sources of heterogeneity:
\textit{Acquisition heterogeneity:} Differences in MRI acquisition protocols, PET radiotracers, scanner hardware, and sequence naming conventions introduce site-specific variability that hampers multi-center model training by introducing non-biological biases. 
\textit{Cohort heterogeneity:} Demographic disparities and uneven distributions of disease stages across cohorts cause significant distributional shifts in both clinical and imaging data~\cite{habes2020disentangling}. 
\textit{Modality-coverage heterogeneity:} Real-world datasets frequently lack complete multimodal coverage (e.g., missing PET scans due to cost or incomplete clinical assessments), preventing models from learning consistent cross-modal relationships.
These factors cause conventional models to overfit to site-specific or modality-specific biases, resulting in poor \jc{cross-center} generalization. 

%这里需要修改一下，简单介绍一下我们的想法是如何去解决上述提到的问题的简单intuition：通过训练的时候注入额外的meta信息知识以做增强模型对这几个方面
To address the above challenges in data heterogeneity for scale-up training, we propose a Collaborative Meta Knowledge Enhancement (COME) framework for dementia etiology diagnosis.
Intuitively, we hypothesize that injecting acquisition semantics, source identifiers, and modality indicators as \jc{prior guidance can promote} the model to distinguish meaningful disease-related patterns from site- or modality-specific artifacts.
Crucially, to prevent \jc{spurious correlations from being amplified in the guidance injection}, we anchor the model optimization within established clinical diagnostic principles, enabling the model to mediate the heterogeneity of multi-source dementia-related data while preserving diagnostic integrity. 
The main contributions are summarized as follows: 
\begin{itemize}
%
% \item We propose a novel Meta Knowledge Enhancement framework that injects acquisition semantics, data source identifiers, and modality indicators into a unified Transformer model. This design enables effective learning from highly heterogeneous multi-center data by explicitly encoding sources of variation that are usually ignored. 
\item We propose a novel Meta Knowledge Enhancement framework that \rev{represents acquisition semantics, data source identifiers, and modality indicators as heterogeneity-aware tokens and injects them into a unified Transformer model through attention-based refinement}. This \rev{token-level modulation} enables effective learning from highly heterogeneous multi-center data by explicitly \rev{modeling} sources of variation that are usually ignored.
\item We develop a trust-region constrained optimization scheme with a reference model to regularize \jc{guidance injection}. This ensures the model to leverage meaningful \jc{meta knowledge to flexibly combat the multi-center data heterogeneity while preventing the model from} capturing spurious correlations, thereby improving robustness. 
\item Our method achieves state-of-the-art performance in dementia classification across seven cohorts, outperforming baselines in both in-domain and challenging cross-center/sequence settings, with strong robustness to distribution shifts. \jc{Comprehensive verification confirms the effectiveness and alignment with the key biomarkers.}
% (e.g., higher Alzheimer’s risk for amyloid/tau-positive cases).

\end{itemize}

\begin{figure*}[t!]
\centerline{\includegraphics[width=0.85\linewidth]{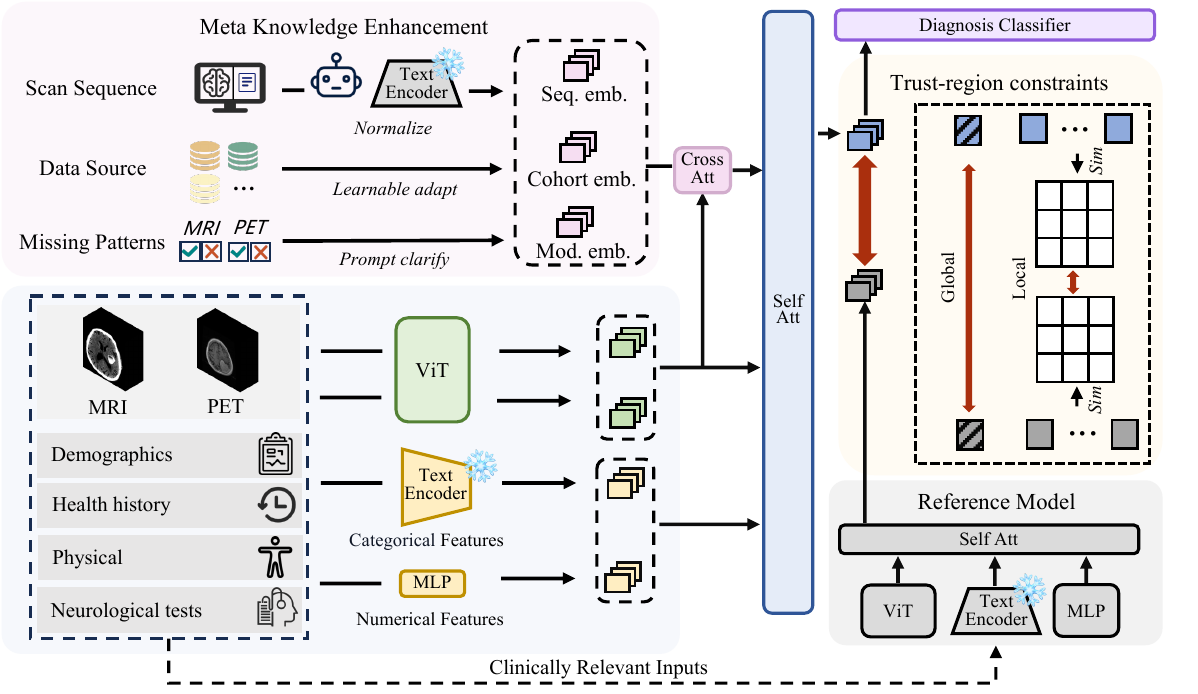}}
\caption{
% Collaborative Meta Knowledge Enhancement framework for multi-center dementia diagnosis. MRI and PET data are encoded using the ViT model, and clinical data is divided into categorical and numerical features, which are encoded and processed separately.
% Integrates acquisition sequence semantics, data source identifiers, and modality coverage embeddings into the diagnosis model, with trust-region constraints ensuring clinical relevance through global feature and local patch alignment. 
\rev{Overview of the Collaborative Meta Knowledge Enhancement (COME) framework for multi-center dementia diagnosis. MRI/PET images and clinical variables are encoded separately and fused in a unified Transformer, while acquisition-sequence, cohort-source, and modality-coverage knowledge are encoded as meta-knowledge embeddings to enhance multimodal representations. A reference model without meta-knowledge enhancement provides global feature and local patch alignment constraints during the constrained optimization phase.}
}

\label{fig:main_method}
\vspace{-12pt}
\end{figure*}

%% file: body/related_works.tex
\section{related works}

\subsection{AI-Assisted Dementia Etiology Diagnosis}
Artificial intelligence has revolutionized dementia diagnosis by enabling automated, accurate, and scalable analysis of complex medical data. Early efforts primarily focused on unimodal approaches: Nawaz et al.~\cite{nawaz2020deep} pioneered deep learning-based classification of Alzheimer’s disease (AD) using structural MRI, while Roy et al.~\cite{roy2019segmentation} demonstrated robust tumor delineation in multi-sequence MRI through segmentation-focused models. Concurrently, machine learning approaches identified potential CSF biomarkers (e.g., TNFRSF4, TGF-$\beta$1) for predicting cognitive decline~\cite{gogishvili2023discovery}. 
However, despite their technical merits, such single-modality approaches exhibit inherent limitations in clinical applicability due to incomplete characterization of disease pathology.
Recent advances have shifted toward multimodal integration and explainability to address these limitations. For example, recent frameworks~\cite{qiu2022multimodal,xue2024ai} have combined clinical, neuroimaging, and biomarker data to achieve high accuracy in differentiating multiple dementia subtypes, supported by interpretability analyses and multi-level validation involving clinical indicators and biomarkers.
While these studies demonstrate the promise of AI for dementia diagnosis, they primarily evaluate within single-cohort or harmonized datasets and rarely address the intrinsic challenges \jc{when scaling up to multi-center data}. In particular, they pay little attention to the pronounced imaging heterogeneity or distribution shifts encountered in cross-center testing, which hinder the direct translation of current AI models into robust, generalizable clinical tools, motivating the need for approaches that explicitly address such real-world variability.

\subsection{\rev{Multi-Task and Heterogeneity-Aware Learning}}
Multi-Task Learning (MTL) has been applied in medical imaging and dementia research to leverage shared structure across related prediction tasks~\cite{tian2022mri,liu2024multi,karani2018lifelong, beljaards2020cross, zhang2024foundation, hashimoto2024multimodal}. In principle, MTL can improve generalization in multi-center studies by exploiting complementary information across tasks.
MTL can be broadly categorized into three types: hard parameter sharing (HPS)~\cite{caruana1993multitask}, soft parameter sharing (SPS)~\cite{duong2015low}, and Mixture-of-Experts (MoE)~\cite{shazeer2017outrageously}. 
HPS approaches typically share feature extractors across all tasks~\cite{caruana1993multitask} and may incorporate multi-objective optimization to balance task-specific losses~\cite{long2017learning,senushkin2023independent}. 
While these methods have been used in AD diagnosis and cognitive assessment prediction~\cite{tian2022mri,liu2024multi}, they often struggle in the presence of strong inter-cohort heterogeneity: HPS tends to overfit toward tasks or centers with dominant signals.
SPS and MoE provide more flexible mechanisms to model inter-task relationships. SPS allows dynamic feature sharing across tasks~\cite{duong2015low,wallingford2022task}, while MoE dynamically routes parameters via learnable gates~\cite{ma2018modeling, hazimeh2021dselect}.
Both strategies have been applied in medical imaging to improve data efficiency and generalization in real-world, multi-center settings~\cite{karani2018lifelong, beljaards2020cross, zhang2024foundation, hashimoto2024multimodal}.
\rev{Beyond conventional MTL, several related lines address heterogeneity from complementary perspectives. Knowledge distillation and model reprogramming transfer information across heterogeneous model or data spaces~\cite{li2024lorkd,zhou2024reprogramming}; disentanglement separates disease-relevant factors from cohort-related variations~\cite{habes2020disentangling}; domain-shift robustness reduces reliance on site-specific or spurious features~\cite{saab2022reducing}; and missing-aware or heterogeneous-data modeling considers incomplete modality coverage and source-dependent variations~\cite{yu2024smart,zhou2024exploring}. These studies suggest that heterogeneity-related information should be systematically considered rather than treated as unobserved noise.}
\rev{Despite these advances, existing frameworks} implicitly learn from data source variation without distinguishing meaningful biological signals from site- or modality-specific information. 
This limits their ability to handle the acquisition, cohort, and modality-coverage heterogeneities common in multi-center dementia datasets, motivating the need for strategies that can explicitly incorporate such meta-knowledge into the learning process.

%% file: body/method.tex
\section{Method}
\subsection{Preliminary}
%介绍一下
\subsubsection{Notation}
We consider a multi-center multi-modal dataset $\{(\mathrm{x}_i, y_i)\}_{i=1}^N$ with $N$ subjects. Each sample $\mathrm{x}_i = \{x_i^m\}_{m=1}^M$ consists of up to $M=3$ modalities: structural MRI ($x_i^{\text{MRI}}$), PET ($x_i^{\text{PET}}$), and tabular clinical data ($x_i^{\text{cli}}$), where missing modalities are indicated by a binary mask $\delta_i \in \{0,1\}^M$ with $\delta_i^m = 1$ if modality $m$ is present. Clinical data are split into categorical $x^{\text{cli}}_{i,\text{cat}}$ and numerical $x^{\text{cli}}_{i,\text{num}}$ features, with the former encoded via a pretrained text encoder and the latter normalized and used as a numeric vector. 
For each subject $i$, every medical image is associated with an acquisition sequence name ~$q_i^m$ ($m \in \{\mathrm{MRI}, \mathrm{PET}\}$), denoting the contrast type (e.g., T1-, T2-weighted) for MRI or radiotracer for PET, along with key post-processing steps. 
The label $y_i$ represents a multi-label diagnosis vector indicating the presence of one or more dementia types. Each sample originates from one of $S$ data collection sites, denoted by source identifier $s_i \in \{1,\dots,S\}$. The goal is to train a multimodal diagnostic model $f(x_i^{\text{MRI}}, x_i^{\text{PET}}, x_i^{\text{cli}})$ that can accurately predict $y_i$ to assist doctors in diagnosis.

\subsubsection{Motivation}
Training robust dementia etiology diagnosis models on multi-center datasets is fundamentally challenged by the pervasive heterogeneity inherent in real-world data. As highlighted in the Introduction, this heterogeneity manifests primarily in three critical dimensions:

\paragraph{Acquisition heterogeneity} Differences in acquisition protocols (e.g., scanner vendors, sequence parameters) and inconsistent naming conventions introduce significant non-biological variations into the imaging data across centers.

\paragraph{Cohort heterogeneity} Differences in disease stages and demographic characteristics lead to diversity in medical imaging and clinical indicators, particularly in dementia-related brain disorders~\cite{habes2020disentangling}. This heterogeneity introduces substantial data distribution shifts during model training.

\paragraph{Modality-coverage heterogeneity} Real-world constraints (cost, availability, patient factors) result in highly variable and often incomplete coverage of multimodal data (e.g., missing MRI or PET scans) across different centers, complicating multimodal fusion and creating data bias.

These compounding factors severely hinder conventional AI models' generalization across unseen centers and handling of incomplete data.
Standard AI approaches, while technically advanced, remain brittle under real-world distribution shifts.
A natural remedy is MTL, which leverages complementarities across centers. However, existing MTL strategies mainly rely on implicit task relationships without explicitly modeling acquisition, cohort, or modality-coverage heterogeneities.
To address these issues, we propose a Collaborative Meta Knowledge Enhancement (COME) framework to inject meta knowledge (acquisition sequences, data source, etc.) into the model via knowledge-enhanced embeddings, enabling adaptive feature customization and improving robustness under pronounced data heterogeneity.
Our framework is illustrated in Fig.~\ref{fig:main_method}, and in the following, we discuss the key components in detail.

\subsection{Collaborative Meta Knowledge Enhancement}
\jc{Our COME framework is designed to inject the meta knowledge into a unified Transformer model to customize the training under the multi-center data heterogeneity}. 
% This enables the ViT encoder to dynamically adapt to heterogeneous medical images of varying quality, source, and modality, while collaboratively fusing image features with clinical data for robust dementia etiology diagnosis.
%
Specifically, different types of meta knowledge are encoded \jc{as knowledge-enhanced embeddings}, which are injected via cross-attention in the early layers to refine feature extraction. 
\jc{Subsequently, the resulting image features are fused with clinical features through self-attention, which are finally fed into a classifier to obtain the diagnosis}.

\subsubsection{Meta Knowledge Encoding Mechanism}
We \jc{consider three types of meta knowledge to combat the heterogeneity}:
\paragraph{Acquisition-aware embeddings}
Different imaging modalities (and even different acquisition sequences of the same modality) exhibit distinct visual characteristics due to varying imaging principles. We hypothesize that incorporating such acquisition knowledge will enable customized processing of heterogeneous imaging inputs.
Due to sequence naming discrepancies across multi-center datasets, we employ a large language model \jc{$f_{\text{std}}(\cdot)$} (DeepSeek~V3~\cite{liu2024deepseek}) \rev{as a prompt-based offline standardizer} to generate standardized technical descriptions $t^m$ from raw sequence names $q^m$, \jc{where $m$ is modality type}. 
\rev{This choice avoids training an additional sequence-name normalizer while providing sufficient instruction-following ability to convert heterogeneous acquisition abbreviations into consistent technical descriptions. Radiologists further checked the LLM-generated descriptions for all unique sequence names as an offline quality-control step and did not identify severe errors in the acquisition-related information.} 
% For example, the raw sequence name: \texttt{T2\_in\_T1-anatomical\_space} is transformed into the standardized description: \texttt{T2-weighted contrast in T1 anatomical space for combined structural and pathological assessment with precise spatial alignment.}
%
% \jc{Note that,} these descriptions are reviewed by the radiologists to ensure clinical \jc{correctness} before being 
\rev{Additional standardization details are summarized in Appendix~\ref{app:seq_standardization}.}
The resulting descriptions are then encoded into textual embeddings $\mathbf{P}_{\text{seq}}^m \in \mathbb{R}^{L_{\text{seq}} \times d}$ via \jc{the pretrained ClinicalBERT encoder $f_{\text{bert}}(\cdot)$}~\cite{liu2025generalist}, where $L_{\text{seq}}$ is the token length and $d$ is the embedding dimension. \jc{We formalize this process as follows:}
$$\mathbf{P}_{\text{seq}}^{m}=f_{\text{bert}}(f_{\text{std}}(q^m)),\quad m\in\{\text{MRI},\text{PET}\}.$$
$\mathbf{P}_{\text{seq}}^{m,i}$ helps inform the model about the characteristics of the imaging sequence, promoting the encoder to quickly adapt to the expected intensity patterns and contrast of that sequence.
\paragraph{Cohort-aware embeddings}
Different data collection sites or cohorts can induce the distribution shifts (due to population demographics, scanner hardware, etc.). Inspired by~\cite{saito2023prefix} \jc{that shows the source representation can implicitly characterize the knowledge of distribution}, we assign a trainable embedding $\mathbf{P}_{\text{cohort}}^{s} \in \mathbb{R}^{d}$ for each source $s$. 
During training, these embeddings are used to capture the distribution biases of the sources, \jc{leaving the backbone model to focus on the generalizable patterns} across diverse data sources.

\paragraph{Modality-coverage embeddings} 
\jc{For the heterogeneity induced by the modality-missing issue, we design tokens to characterize the modality coverage information. Specifically,} 
for each modality $m$, we maintain two trainable embeddings: $\mathbf{P}^{m}$ (present) and $\mathbf{\bar{P}}^{m}$ (missing). Given an indicator $\delta\in\{0,1\}$, \jc{the modality-coverage embedding can be expressed as $\mathbf{P}_{\text{mod}}^{m} = \delta^{m}\mathbf{P}^{m}+(1-\delta^{m})\mathbf{\bar{P}}^{m} \in \mathbb{R}^{d}$}, which is used to explicitly inform the encoder about remaining modalities, ensuring robust fusion \jc{when encountering missing inputs}.

In summary, we have transformed the meta knowledge for each sample into a collection of knowledge embeddings: 
\begin{equation}
    \jc{\mathbf{P}_{\text{meta}} = \left\{ \mathbf{P}_{\text{seq}}^{\text{MRI}}, \mathbf{P}_{\text{seq}}^{\text{PET}},\mathbf{P}_{\text{mod}}^{\text{MRI}}, \mathbf{P}_{\text{mod}}^{\text{PET}},\mathbf{P}^{s}_{\text{cohort}} \right\}.}
\end{equation}

\subsubsection{Knowledge-enhanced Feature Refinement and Fusion}
The meta knowledge embeddings are strategically injected to enhance image feature extraction and multimodal fusion. We detail how the meta knowledge tokens are injected into the imaging features, followed by the unified feature fusion process.
\paragraph{Refine Image Features with Knowledge Embeddings}
For each present imaging modality $m \in \{\text{MRI}, \text{PET}\}$ of a sample, we extract initial features ${I}^{m}_{0} \in \mathbb{R}^{N_p \times d} $ from $x^m$ using a shared ViT backbone $f_{\text{vit}}(\cdot)$, 
where $N_p$ is the number of image patches,
% then progressively \jc{refine the features with meta knowledge}, applying a $J$-layer cross-attention module:
% %
% \begin{equation}
% \begin{aligned}
%     &\mathbf{I}^{m}_j = \text{CrossAttn}\left(\mathbf{Q}_j, \mathbf{K}_j, \mathbf{V}_j \right)_{:N_p} ~~\text{for} ~ j = 1, \dots, J~, \\
%     &\text{with}\ 
%     \begin{cases}
%         \mathbf{Q}_j = \mathbf{P}^{s}_{\text{cohort}} \\
%         \mathbf{K}_j = \mathbf{I}^{m}_{j-1} \oplus \mathbf{P}_{\text{seq}}^{m} \oplus \mathbf{P}_{\text{mod}}^{m} \\
%         \mathbf{V}_j = \mathbf{I}^{m}_{j-1} \oplus \mathbf{P}_{\text{seq}}^{m} \oplus \mathbf{P}_{\text{mod}}^{m}
%     \end{cases}\text{and}~ \mathbf{I}^{m}_{0} = f_{\text{vit}}(x^m)~,
% \end{aligned}
% \end{equation}
% where $\text{CrossAttn}$ denotes the cross-attention operation, and $ \mathbf{P}^{s}_{\text{cohort}}$ (cohort embeddings) acts as the query to dynamically modulate features based on the data source.
% \rev{Here, $\oplus$ denotes token-wise concatenation along the sequence/token dimension rather than element-wise addition; thus,} 
% the key and value incorporate image embeddings ($\mathbf{I}^m$), acquisition-aware embeddings ($\mathbf{P}_{\text{seq}}^m$), and modality-status embeddings ($\mathbf{P}_{\text{mod}}^m$). This enables the adaptive refinement of image features using acquisition-related knowledge while handling missing modalities. 
We then progressively refine the features with meta knowledge.
\rev{At the $j$-th refinement layer, the current image tokens are first augmented with the acquisition-aware and modality-coverage tokens to form a sequence conditioned on meta knowledge:
\begin{equation}
\begin{aligned}
    \mathbf{G}^{m}_{j-1} &= \mathbf{I}^{m}_{j-1} \oplus \mathbf{P}_{\text{seq}}^{m} \oplus \mathbf{P}_{\text{mod}}^{m},
    \quad \mathbf{I}^{m}_{0} = f_{\text{vit}}(x^m).
\end{aligned}
\end{equation}
The cohort-aware embedding is then used as a query over this sequence to generate a cohort-conditioned token, which is appended before Transformer refinement:
\begin{equation}
\begin{aligned}
    \mathbf{C}^{m}_{j} &=
    \operatorname{Attn}_{j}^{c}
    \left(
    \mathbf{P}^{s}_{\text{cohort}},
    \mathbf{G}^{m}_{j-1},
    \mathbf{G}^{m}_{j-1}
    \right),\\
    \mathbf{H}^{m}_{j} &=
    \operatorname{Trans}_{j}
    \left(
    \mathbf{G}^{m}_{j-1} \oplus \mathbf{C}^{m}_{j}
    \right), \\
    \mathbf{I}^{m}_{j} &= \mathbf{H}^{m}_{j}[:N_p], \quad j=1,\dots,J.
\end{aligned}
\end{equation}
Here, $\oplus$ denotes token-wise concatenation along the token dimension, and $\mathbf{H}^{m}_{j}[:N_p]$ keeps the updated image-token segment after refinement. Projection matrices inside $\operatorname{Attn}_{j}^{c}$ and $\operatorname{Trans}_{j}$ are omitted from the equations for compactness. 
}
This \rev{attention-based refinement} enables \rev{image features to be adaptively modulated by} acquisition-related knowledge while handling missing modalities. 
\rev{We set $J=2$ to enable iterative meta-knowledge refinement in the main experiments.}
The refined features $\mathbf{I}^{m}_J$ retain only the image feature part (first $N_p$ tokens) and then are combined via an average over available modalities to produce a unified imaging embedding $\mathbf{U}$, where the average includes only modalities present in the sample, as indicated by the binary mask $\delta^m$:
\begin{equation}
    \mathbf{U} = \frac{1}{\sum_{m} \delta^m} \sum_{m \,\in \, \{m \mid \delta^m = 1\}} \mathbf{I}^{m}_J
\end{equation}

\paragraph{Unified Feature Fusion}
To enhance diagnostic accuracy, our model also incorporates clinical data such as demographics, health history, physical metrics, and neurological test scores. These clinical data are categorized into two types: categorical features and numerical features. 
\rev{For categorical variables, available fields are concatenated as short clinical statements, while missing fields are omitted. For numerical variables, missing entries are represented by zero-valued inputs together with observation masks before MLP projection.}
Categorical features $x^{\text{cli}}_{\text{cat}}$ are encoded by the same ClinicalBERT encoder $f_\text{bert}(\cdot)$, while numerical features $x^{\text{cli}}_{\text{num}}$ are projected via an MLP $f_\text{mlp}(\cdot)$. Then, we linearly transform these features as clinical embedding $\mathbf{t}$ by $\mathbf{W}_c$, which are fused with the refined image features $\mathbf{U}$ as the final representation via self-attention layers,  
\rev{where $\mathbf{U} \oplus \mathbf{t}$ means appending the clinical embedding token to the refined imaging token sequence.} 
Formally, we summarize this process by the following equation:
\begin{equation} \label{eq:rep}
\begin{split}
    & \quad\quad \mathbf{z} = \text{SelfAttn}\left( \left[ \mathbf{U} \oplus \mathbf{t} \right] \right)\in \mathbb{R}^{N \times d}, \\
    & \jc{\text{where } \mathbf{t} = \mathbf{W}_c [f_{\text{bert}}(x^{\text{cli}}_{\text{cat}}); f_{\text{mlp}}(x^{\text{cli}}_{\text{num}})] \in \mathbb{R}^d,}
\end{split}
\end{equation}
\jc{and $N$ is the patch token number.} The fused representation $z$ is pooled via attention by $\text{AttPool}(\cdot)$ and fed into an MLP head $\phi_{\text{mlp}}(\cdot)$ to predict multi-label dementia  $\hat{\mathbf{y}}$, with the classification loss $\mathcal{L}_{\text{cls}}$ computed by binary cross-entropy (BCE): 
\begin{equation}
\hat{\mathbf{y}} = \phi_{\text{mlp}}(\text{AttPool}(\mathbf{z})), \quad \mathcal{L}_{\text{cls}} = \ell_{\text{BCE}}(\hat{\mathbf{y}}, \mathbf{y}).
\end{equation}

\begin{figure}[t!]
\centering
\includegraphics[width=0.42\textwidth]{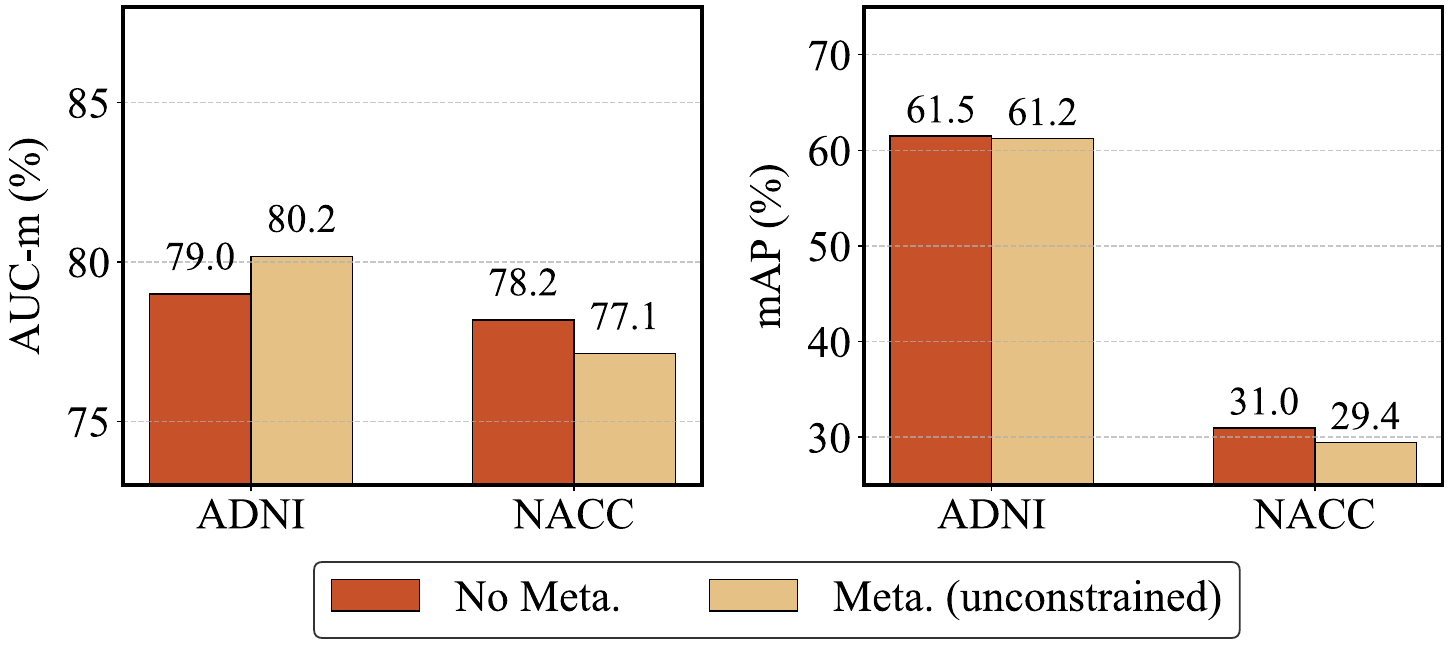} %
\caption{Performance with and without meta knowledge in cross-center evaluation. The comparison includes: (1) standard training without meta knowledge, and (2) direct (unconstrained) integration of meta knowledge. Left: AUC-m; Right: mAP. ADNI/NACC as test sets, other centers' data for training.}
\label{fig:method_intuition}
\vspace{-12pt}
\end{figure}
\subsection{Trust-region Constrained Optimization}
Meta knowledge enhances models with task-relevant complementary information \jc{characterizing} the heterogeneous inputs. 
However, unconstrained integration may introduce potential risks by encouraging spurious correlations: missing modality patterns (e.g., "Missing PET = Healthy") create non-causal diagnostic shortcuts~\cite{yu2024smart}, while neural networks over-rely on dataset-specific image artifacts (e.g., acquisition-related features) as predictors~\cite{saab2022reducing}.
Such shortcuts \jc{mis-link} non-generalizable artifacts rather than clinically meaningful biomarkers, causing severe generalization failures on external cohorts. As illustrated in Fig.~\ref{fig:method_intuition}, compared to standard multi-modal model training, the gains from unconstrained injection of meta knowledge are limited and often lead to adverse effects. 

To mitigate this risk, we propose a trust-region inspired optimization to confine meta-knowledge integration within a clinically valid trust-region.
The trust-region is defined as the space of representations learned exclusively from core clinical inputs (imaging and clinical data), free from heterogeneity-induced biases.
%
% Inspired by the hard update mechanism in constrained learning~\cite{gorbatovski2024learn}, we instantiate the trust-region optimization via a reference model (identical architecture to the main model but \jc{without meta knowledge enhancement}) that dynamically maintains the trust-region boundary through hard parameter updates every $k$ training epochs:
% $$
% \theta_{\text{ref}} \leftarrow \theta_{\text{main}} \quad \text{when}\quad t\mod k = 0
% $$
% where $\theta_{\text{main}}$ and $\theta_{\text{ref}}$ denote the parameters of the main model and the reference model respectively, and $t$ denotes the training step.
% Critically, this reference model continuously captures the trust region of clinically reliable features, providing a stable target to constrain the main model's meta knowledge enhanced representations.
%
Inspired by the hard update mechanism in constrained learning, we instantiate the trust-region optimization via a reference model (identical architecture to the main model but without meta knowledge enhancement) that \rev{serves as an intermittently refreshed trust-region anchor through} hard parameter updates every $k$ training epochs:
$$
\theta_{\text{ref}} \leftarrow \theta_{\text{main}} \quad \text{when}\quad t\mod k = 0
$$
where $\theta_{\text{main}}$ and $\theta_{\text{ref}}$ denote the parameters of the main model and the reference model respectively, and \rev{$t$ denotes the epoch index.}
Critically, this reference model \rev{is not optimized independently between hard updates; instead, it remains frozen and provides} a stable target to constrain the main model's meta knowledge enhanced representations.
To \jc{properly constrain the model optimization by the reference model} at both macroscopic diagnostic and microscopic pathological levels, which is critical for dementia etiology~\cite{jack2018nia}, we design a dual-level alignment: 1) Global feature alignment stabilizes macroscopic reasoning against spurious correlations with data sources/modality missing patterns; 
2) Local patch alignment ensures focus on subtle lesion regions rather than acquisition-sequence artifacts.
\rev{Accordingly, we use a minibatch contrastive objective~\cite{oord2018representation} for global alignment to compare matched main/reference patient-level representations against other samples in the batch, while using cosine-affinity MSE for local alignment to preserve within-sample patch structure without introducing unreliable patch-level negatives.}

\textbf{Global Feature Alignment}: 
%下标i误用，前面是sample index 这里是batch index
% To prevent the meta knowledge enhancement from steering the main model's high-level semantic representations too far from the diagnostically reliable features learned from the core data, we enforce global feature alignment. 
\jc{This global alignment is to} prevent the meta knowledge enhancement from steering the main model's high-level semantic representations too far from the diagnostically reliable features learned from the core data.
%
% \jcr{The following notations should reuse previous designs.}
Let \jc{$\hat{\mathbf{z}}$} denote the representation of the reference model after the SelfAttn layer \jc{(and $\mathbf{z}$ has been clarified in Eq.~\eqref{eq:rep}). We compute the mean-pooled embedding of all attention tokens in $\hat{\mathbf{z}}$ and $\mathbf{z}$, denoted as $\hat{\mathbf{z}}^G$ and $\mathbf{z}^G$ respectively}. We then construct a symmetric contrastive loss~\cite{oord2018representation} to align matched pairs $(\hat{\mathbf{z}}_i^G, \mathbf{z}_i^G)$ from the same patient while minimizing the similarity with others in a training minibatch as follows
% \begin{align}
% \begin{split}
% \mathcal{L}_g = \frac{1}{2} \Bigg[ 
% \frac{\exp\left(\frac{\langle \hat{\mathbf{z}}_{i}^G, \mathbf{z}_{i}^G \rangle}{\tau}\right)}{\sum_{b=1}^{|B|} \exp\left(\frac{\langle \hat{\mathbf{z}}_{i}^G, \mathbf{z}_{b}^G \rangle}{\tau}\right)}
% + \frac{\exp\left(\frac{\langle \mathbf{z}_{i}^G, \hat{\mathbf{z}}_{i}^G \rangle}{\tau}\right)}{\sum_{b=1}^{|B|} \exp\left(\frac{\langle \mathbf{z}_{i}^G, \hat{\mathbf{z}}_{b}^G \rangle}{\tau}\right)}
% \Bigg],
% \end{split}
% \label{L_g}
% \end{align}
% where $\langle \cdot,\cdot \rangle$ denotes cosine similarity

% \begin{align}
% \begin{split}
% \mathcal{L}_g = -\frac{1}{2} \Bigg[ 
% \frac{\exp\left(\frac{\rev{\operatorname{sim}(\hat{\mathbf{z}}_{i}^G, \mathbf{z}_{i}^G)}}{\tau}\right)}{\sum_{b=1}^{|B|} \exp\left(\frac{\rev{\operatorname{sim}(\hat{\mathbf{z}}_{i}^G, \mathbf{z}_{b}^G)}}{\tau}\right)}
% + \frac{\exp\left(\frac{\rev{\operatorname{sim}(\mathbf{z}_{i}^G, \hat{\mathbf{z}}_{i}^G)}}{\tau}\right)}{\sum_{b=1}^{|B|} \exp\left(\frac{\rev{\operatorname{sim}(\mathbf{z}_{i}^G, \hat{\mathbf{z}}_{b}^G)}}{\tau}\right)}
% \Bigg],
% \end{split}
% \label{L_g}
% \end{align}
\begin{align}
\begin{split}
\mathcal{L}_g
= -\frac{1}{2|B|}\sum_{i=1}^{|B|} \Bigg[
&\log
\frac{\exp\left(\rev{\operatorname{sim}(\hat{\mathbf{z}}_{i}^G, \mathbf{z}_{i}^G)}/\tau\right)}
{\sum_{b=1}^{|B|} \exp\left(\rev{\operatorname{sim}(\hat{\mathbf{z}}_{i}^G, \mathbf{z}_{b}^G)}/\tau\right)}
\\
&+ \log
\frac{\exp\left(\rev{\operatorname{sim}(\mathbf{z}_{i}^G, \hat{\mathbf{z}}_{i}^G)}/\tau\right)}
{\sum_{b=1}^{|B|} \exp\left(\rev{\operatorname{sim}(\mathbf{z}_{i}^G, \hat{\mathbf{z}}_{b}^G)}/\tau\right)}
\Bigg],
\end{split}
\label{L_g}
\end{align}
where \rev{$\operatorname{sim}(\mathbf{a},\mathbf{b})=\frac{\mathbf{a}\cdot\mathbf{b}}{\|\mathbf{a}\|_2\|\mathbf{b}\|_2}$ denotes cosine similarity,} $\tau$ is a temperature and $|B|$ is the minibatch size and the subscript $i$ denotes the patient index within the batch.
This loss regularizes the global embedding space of the main model to remain close to that of the reference model, ensuring that the integration of meta knowledge does not \jc{destroy the overall semantic direction}. 

\textbf{Local Patch Alignment}: \jc{This local alignment builds the patch-level constraint, emphasizing the consistency in fine-grained pathological patterns.}
It aligns the structural relationships between patch tokens of the main model ($\mathbf{z}$) and those of the reference model ($\hat{\mathbf{z}}$).
Specifically, we compute an attention-weighted representation $\tilde{\mathbf{z}}_i$ by using $\hat{\mathbf{z}}$  as queries on $\mathbf{z}$:
\begin{align}
\tilde{\mathbf{z}} = \mathrm{softmax}\Big(\frac{\hat{\mathbf{z}} \mathbf{z}^\top}{\sqrt{d}}\Big) \mathbf{z} \in 
\mathbb{R}^{N \times d},\end{align}
where each row of $\tilde{\mathbf{z}}$ is a weighted combination of the main model patch tokens in $\mathbf{z}$, with weights proportional to the patch similarity to the corresponding token in $\hat{\mathbf{z}}$. 
Next, we obtain the self-similarity matrices for  $\hat{\mathbf{z}}$ and $\tilde{\mathbf{z}}$:
% \begin{align}
% R_{ij}(\hat{\mathbf{z}}) = \frac{ \hat{\mathbf{z}}_i \cdot \hat{\mathbf{z}}_j }{ \|\hat{\mathbf{z}}_i\|_2 \cdot \|\hat{\mathbf{z}}_j\|_2 }, \quad \forall i,j = 1,\dots,N
% \label{eq:R}
% \end{align}
\begin{align}
R_{ij}(\hat{\mathbf{z}}) = \rev{\operatorname{sim}(\hat{\mathbf{z}}_i,\hat{\mathbf{z}}_j)}, \quad \forall i,j = 1,\dots,N
\label{eq:R}
\end{align}
and $R_{ij}(\tilde{\mathbf{z}})$ is computed analogously by replacing $\hat{\mathbf{z}}$ with $\tilde{\mathbf{z}},$ in Eq.~\eqref{eq:R}. Here $R(\cdot)$ captures the normalized patch-wise affinity (similarity) structure of the token set. We then minimize the discrepancy between these affinity matrices using a mean squared error (MSE) loss:
\begin{align}
\mathcal{L}_l = \frac{1}{N^2} \sum_{i,j=1}^{N} \Big(R_{ij}(\hat{\mathbf{z}}) - R_{ij}(\tilde{\mathbf{z}})\Big)^2,
\label{L_l}
\end{align}
which encourages the main model to focus on the same sub-region features as the reference model, thereby reinforcing attention on disease-related regions 
% (e.g., cortical atrophy patterns or pathological protein depositions) 
while mitigating distraction from spurious correlations between acquisition-specific characteristics and diagnostic labels.

\subsection{Training Process}
We implement a two-phase training procedure \jc{regarding our COME framework}. In the initialization phase, \rev{to establish a diagnosis-oriented anchor before meta-knowledge enhancement,} the main model is trained solely with the classification loss $\mathcal{L}_{\text{cls}}$ using only imaging and clinical data as input, without incorporating meta-knowledge embeddings.
\rev{After this phase, the reference model is initialized by copying the main-model parameters, i.e., $\theta_{\text{ref}}\leftarrow\theta_{\text{main}}$.}
In the constrained optimization phase, we activate trust-region constraints by introducing a reference model that is updated periodically via hard parameter copying. 
\rev{For each training batch, the meta-knowledge-enhanced main model produces the fused representation $\mathbf{z}$ and prediction $\hat{\mathbf{y}}$, while the reference model processes the same inputs without meta-knowledge enhancement to produce $\hat{\mathbf{z}}$. The reference model is kept fixed between hard updates; its parameters are refreshed by $\theta_{\text{ref}}\leftarrow\theta_{\text{main}}$ every $k$ epochs, with $k=2$ in the main experiments. The two representations are used to compute the global and local alignment losses.} 
The main model, enhanced with meta knowledge, is optimized jointly with the classification loss $\mathcal{L}_{\text{cls}}$, global feature alignment loss $\mathcal{L}_g$, and the local patch alignment loss $\mathcal{L}_l$ as:
 \begin{align}
\mathcal{L} = \mathcal{L}_{\text{cls}} + \lambda(\mathcal{L}_g + \mathcal{L}_l),
\end{align}
where $\lambda=0.2$ balances constraint strength. 
\rev{Together, the initialization phase, constrained optimization phase, and periodic reference-model update define the overall training workflow of COME.}
% The full training pipeline is summarized in Algorithm~\ref{alg:training}.

%% file: body/exp.tex
\section{Experiments}
\subsection{Datasets}
To simulate the heterogeneity inherent in multi-center data, we integrated seven distinct datasets: the National Alzheimer’s Coordinating Center (NACC)~\cite{beekly2004national}, the Alzheimer’s Disease Neuroimaging Initiative (ADNI)~\cite{mueller2005ways}, the Australian Imaging, Biomarker and Lifestyle Flagship Study of Ageing (AIBL)~\cite{ellis2006australian}, the Parkinson’s Progression Marker Initiative (PPMI)~\cite{marek2011parkinson}, the Open Access Series of Imaging Studies-4 (OASIS)~\cite{marcus2010open}, the Frontotemporal Lobar Degeneration Neuroimaging Initiative (NIFD)~\cite{boxer2013frontotemporal}, and the 4 Repeat Tauopathy Neuroimaging Initiative (4RTNI)~\cite{dutt2016progression}. These datasets collectively encompass multimodal clinical, structural MRI (acquired under varying protocols), and PET imaging data (utilizing diverse radiotracers).
Following the processing protocol in \cite{xue2024ai}, 
% we standardized the dementia etiology labels in these datasets.
\rev{we derived diagnostic labels from the cohort-provided clinical diagnosis fields. Participants were categorized as NC, MCI, or DE, and DE cases were further annotated with etiology labels, including AD, LBD, VD, FTD, and PSY. The definitions of these harmonized diagnostic labels are summarized in Table~\ref{tab:etiology_definitions}. Labels were assigned only when the corresponding etiology was officially documented in a cohort and could be mapped to the harmonized schema.  Etiology labels were allowed to be non-mutually exclusive to preserve documented mixed dementia.}
%
% The final samples comprised 5119 participants with normal cognition (NC), 2283 with mild cognitive impairment (MCI), and 2231 with dementia (DE). Among the dementia cases, 1778 had Alzheimer’s disease dementia (AD), 119 Lewy body dementia (LBD), 115 vascular dementia (VD), 357 frontotemporal dementia (FTD), and 73 psychiatric disease-associated dementia (PSY).  
\rev{Dataset-specific diagnosis sources used for label harmonization are summarized in Appendix~\ref{app:label_harmonization}.}

The final samples comprised 5119 NC, 2283 MCI, and 2231 DE participants. Among the DE cases, the etiology-label counts were 1778 AD, 119 LBD, 115 VD, 357 FTD, and 73 PSY.  
\rev{Among the 2231 dementia-etiology cases, 2036 (91.26\%) had one etiology label and 195 (8.74\%) had two or more etiology labels. Mixed dementia was therefore represented as a multi-label target.}
% In our experiment, each participant contributed up to one MRI and one PET scan, selected from the most recent acquisition. Clinical data and diagnoses were similarly drawn from this assessment. 
\rev{Each participant appears only once in the final dataset. Specifically, each participant contributed up to one MRI and one PET scan selected from the most recent eligible acquisition, and the clinical variables and diagnoses were drawn from the same assessment. This person-level selection avoids repeated-measure leakage across splits and keeps the imaging, clinical variables, and diagnostic labels temporally aligned.}
\rev{For acquisition-sequence standardization, the seven cohorts contained 247 unique raw sequence names in total. Radiologists reviewed the LLM-generated descriptions for these unique names as an offline quality-control step; no outputs required manual correction, and no sequence names were discarded as unstandardizable.}
Detailed information for each dataset is summarized in Table~\ref{tab:datainfo}.
\rev{For the clinical-variable input, we harmonized 43 numerical and 9 categorical variables across cohorts. Because the cohorts used different assessment forms, variable coverage was incomplete; the corresponding missingness statistics are summarized in Table~\ref{tab:clinical_missingness}.}

\begin{table}[tb]
  \centering

  \caption{\rev{Definitions of the harmonized diagnostic labels.}}
  \label{tab:etiology_definitions}
  \renewcommand{\arraystretch}{1.45}
  \setlength{\tabcolsep}{8pt}
   \resizebox{0.47\textwidth}{!}{
   \rev{
  \begin{tabularx}{\linewidth}{l X}
    \toprule[2pt]
    Label & Definition \\
    \midrule[1.2pt]
    NC  & Normal cognition. \\
    MCI & Mild cognitive impairment. \\
    DE  & Dementia. \\
    AD  & Alzheimer's disease dementia. \\
    LBD & Lewy body dementia, including dementia with Lewy bodies and Parkinson's disease dementia. \\
    VD  & Vascular dementia or vascular brain injury, including stroke. \\
    FTD & Frontotemporal lobar degeneration and related variants, including primary progressive aphasia, corticobasal degeneration/syndrome, and progressive supranuclear palsy. \\
    PSY & Psychiatric disease-associated dementia, including depression, schizophrenia, bipolar disorder, anxiety, post-traumatic stress disorder, or other documented psychiatric contributors. \\
    \bottomrule[2pt]
  \end{tabularx}
  }}
  % \vspace{-6pt}
\end{table}

\begin{table}[tb]
  \centering
  \caption{Information of the datasets used in the experiment\rev{, including the number of unique raw sequence names}.}
  \resizebox{0.47\textwidth}{!}{
    \begin{tabular}{l|llll|l}
    \toprule[2pt]
    Dataset & Subj. no. & MRI no. & PET no. & \rev{Seq. no.} & Disease involved \\
    \midrule[1.2pt]
    NACC  & 5086  & 4344  & 2415  & \rev{86} & MCI AD LBD VD FTD PSY \\
    ADNI  & 2569  & 1754  & 2133  & \rev{50} & MCI AD \\
    AIBL  & 846   & 689   & 824   & \rev{34} & MCI AD \\
    PPMI  & 403   & 285   & 125   & \rev{33} & MCI \\
    OASIS & 366   & 366   & -     & \rev{1}  & MCI AD LBD VD FTD \\
    NIFD  & 286   & 286   & 153   & \rev{32} & FTD \\
    4RTNI & 78    & 78    & -     & \rev{11} & MCI FTD \\
    \bottomrule[2pt]
    \end{tabular}
    }
  \label{tab:datainfo}%
  \vspace{-6pt}
\end{table}%

\subsection{Evaluation Metrics} 
Owing to the multi-label nature of dementia etiology diagnosis, in which patients may present multiple co-occurring etiologies, we employ \rev{four} complementary metrics: macro-averaged Area Under the ROC Curve (AUC-m), mean Average Precision (mAP), macro F1-score (F1-m) \rev{and macro-averaged Recall (Recall-m)}.
\cmx{These metrics collectively reduce the impact of class imbalance and heterogeneous label distributions across centers.}
Specifically, AUC-m measures class-wise ranking performance by averaging unweighted AUC values across all classes, ensuring that minority classes like rare dementia subtypes are equally weighted. The mAP calculates the mean of per-class average precision, highlighting accurate detection of positive cases in imbalanced settings. F1-m reports the unweighted mean of per-class F1-scores, balancing precision and recall across all classes regardless of prevalence differences. \rev{Recall-m directly measures class-wise sensitivity to positive etiology labels.}
% \rev{For F1-m and Recall-m, probabilities are binarized with a fixed threshold of 0.5 for all labels.}
% 

\begin{table}[tb]
  \centering
  \caption{\rev{Coverage of the harmonized clinical variables used in this study, including 43 numerical and 9 categorical variables.}}
  \resizebox{0.42\textwidth}{!}{
  \rev{
    \begin{tabular}{l|ccc|c}
    \toprule[2pt]
    \multirow{2}{*}{Dataset} & \multicolumn{3}{c|}{Observed variables (\%)} & \multirow{2}{*}{Median missing} \\
    \cmidrule{2-4}
     & Overall & Numerical & Categorical & \\
    \midrule[1.2pt]
    NACC  & 67.1 & 68.0 & 62.9 & 16 \\
    ADNI  & 74.5 & 72.5 & 83.7 & 13 \\
    AIBL  & 13.5 & 11.6 & 22.2 & 45 \\
    PPMI  & 10.1 & 7.0  & 24.6 & 46 \\
    OASIS & 20.2 & 17.5 & 33.3 & 41 \\
    NIFD  & 18.4 & 17.8 & 21.3 & 42 \\
    4RTNI & 42.0 & 46.3 & 21.5 & 29 \\
    \midrule[1.2pt]
    Overall & 58.5 & 57.0 & 63.7 & 16 \\
    \bottomrule[2pt]
    \end{tabular}}
    }
  \label{tab:clinical_missingness}%
  \vspace{-6pt}
\end{table}%

\subsection{Baseline}
We evaluate COME against two categories of established approaches. For multi-task learning (MTL) baselines addressing data heterogeneity, we compare COME with 1) Single-Task Learning, where separate models are trained for each dataset; 2) Hard Parameter Sharing (HPS)~\cite{long2017learning} and Aligned~\cite{senushkin2023independent}, which represent fundamental shared-representation MTL approaches;
3) Multi-gate Mixture-of-Experts (MMoE)~\cite{ma2018modeling} and DSelect-K~\cite{hazimeh2021dselect}, which handle task conflicts through gated expert networks. 
For medical diagnostic baselines, we consider the general multi-modal frameworks RadNet~\cite{zheng2024large} and M4Survive~\cite{lee2025multi}, adapted to our task, as well as the dementia-specific neuroimaging models NCOMMS22~\cite{qiu2022multimodal} and NMED24~\cite{xue2024ai}. This diverse set of baselines enables a rigorous evaluation of the ability of COME to address multi-center heterogeneity and modality missingness.

\begin{table*}[h!]
  \centering
  \caption{In-domain performance comparison on seven multi-center dementia cohorts. \rev{Results are averaged over five random seeds. Per-dataset columns report the seed mean, and AVG columns report mean $\pm$ standard deviation.} Best results highlighted in bold red, second best underlined.}
  \resizebox{\textwidth}{!}{
    \begin{tabular}{c|cc|cc|cc|cc|cc|cc|cc|cc}
    \toprule[2pt]
          \multirow{2}[1]{*}{Method}& \multicolumn{2}{c|}{PPMI} & \multicolumn{2}{c|}{NIFD} & \multicolumn{2}{c|}{4RTNI} & \multicolumn{2}{c|}{OASIS} & \multicolumn{2}{c|}{AIBL} & \multicolumn{2}{c|}{NACC} & \multicolumn{2}{c|}{ADNI} & \multicolumn{2}{c}{AVG} \\
    \cmidrule{2-17}
 & AUC-m   & mAP   & AUC-m   & mAP   & AUC-m   & mAP   & AUC-m   & mAP   & AUC-m   & mAP   & AUC-m   & mAP   & AUC-m   & mAP   & AUC-m   & mAP \\
    \midrule[1.2pt]
    Single\_task & \rev{70.49} & \rev{66.55} & \rev{87.25} & \rev{86.49} & \rev{48.31} & \rev{46.31} & \rev{61.86} & \rev{\underline{31.88}} & \rev{88.67} & \rev{\underline{76.89}} & \rev{87.16} & \rev{36.59} & \rev{86.29} & \rev{73.68} & \rev{75.72 $\pm$ 0.66} & \rev{59.77 $\pm$ 0.33} \\
    \midrule[1.2pt]
    HPS & \rev{78.78} & \rev{56.34} & \rev{95.79} & \rev{95.89} & \rev{55.12} & \rev{42.16} & \rev{62.69} & \rev{27.78} & \rev{87.85} & \rev{73.36} & \rev{85.24} & \rev{37.21} & \rev{81.20} & \rev{66.25} & \rev{78.10 $\pm$ 0.15} & \rev{57.00 $\pm$ 0.09} \\
    Aligned & \rev{\underline{86.09}} & \rev{\underline{66.76}} & \rev{96.62} & \rev{96.66} & \rev{68.95} & \rev{56.68} & \rev{59.17} & \rev{25.93} & \rev{88.63} & \rev{75.07} & \rev{87.06} & \rev{39.85} & \rev{82.78} & \rev{69.57} & \rev{\underline{81.33 $\pm$ 0.63}} & \rev{61.50 $\pm$ 0.94} \\
    MMOE & \rev{84.38} & \rev{61.61} & \rev{96.59} & \rev{96.63} & \rev{58.69} & \rev{48.13} & \rev{\underline{68.24}} & \rev{28.25} & \rev{87.46} & \rev{70.91} & \rev{84.86} & \rev{36.43} & \rev{82.15} & \rev{69.18} & \rev{80.34 $\pm$ 0.08} & \rev{58.74 $\pm$ 0.19} \\
    DSelect-K & \rev{78.06} & \rev{65.00} & \rev{95.50} & \rev{95.36} & \rev{61.30} & \rev{52.84} & \rev{63.25} & \rev{26.46} & \rev{\underline{89.70}} & \rev{74.32} & \rev{86.58} & \rev{37.74} & \rev{86.14} & \rev{73.35} & \rev{80.07 $\pm$ 1.69} & \rev{60.73 $\pm$ 1.19} \\
    \midrule[1.2pt]
    
    Radnet & \rev{77.76} & \rev{62.79} & \rev{96.85} & \rev{96.87} & \rev{66.33} & \rev{59.28} & \rev{65.30} & \rev{28.08} & \rev{86.44} & \rev{70.93} & \rev{\underline{88.14}} & \rev{\underline{42.09}} & \rev{86.31} & \rev{\underline{73.77}} & \rev{81.02 $\pm$ 1.84} & \rev{61.97 $\pm$ 1.36} \\
    M4Survive & \rev{84.10} & \rev{63.61} & \rev{96.38} & \rev{96.51} & \rev{\underline{71.69}} & \rev{59.78} & \rev{60.51} & \rev{25.63} & \rev{88.21} & \rev{75.17} & \rev{86.20} & \rev{36.74} & \rev{80.68} & \rev{66.84} & \rev{81.11 $\pm$ 0.88} & \rev{60.61 $\pm$ 1.06} \\
    NCOMMS22 & \rev{74.19} & \rev{59.11} & \rev{96.77} & \rev{96.97} & \rev{69.46} & \rev{\underline{64.85}} & \rev{57.66} & \rev{24.31} & \rev{88.09} & \rev{74.86} & \rev{85.72} & \rev{36.88} & \rev{79.95} & \rev{65.80} & \rev{78.83 $\pm$ 0.28} & \rev{60.40 $\pm$ 0.33} \\
    NMED24 & \rev{70.00} & \rev{61.29} & \rev{\underline{96.93}} & \rev{\underline{97.13}} & \rev{69.13} & \rev{63.95} & \rev{61.50} & \rev{26.04} & \rev{\textcolor{red}{\textbf{90.75}}} & \rev{\textcolor{red}{\textbf{77.62}}} & \rev{86.99} & \rev{36.50} & \rev{\underline{86.45}} & \rev{73.60} & \rev{80.25 $\pm$ 1.41} & \rev{\underline{62.30 $\pm$ 0.74}} \\
    \midrule[1.2pt]
    \rowcolor[rgb]{ .949,  .949,  .949} Ours & \rev{\textcolor{red}{\textbf{92.77}}} & \rev{\textcolor{red}{\textbf{79.40}}} & \rev{\textcolor{red}{\textbf{97.16}}} & \rev{\textcolor{red}{\textbf{97.32}}} & \rev{\textcolor{red}{\textbf{74.96}}} & \rev{\textcolor{red}{\textbf{65.37}}} & \rev{\textcolor{red}{\textbf{69.63}}} & \rev{\textcolor{red}{\textbf{32.27}}} & \rev{89.00} & \rev{74.06} & \rev{\textcolor{red}{\textbf{89.04}}} & \rev{\textcolor{red}{\textbf{42.67}}} & \rev{\textcolor{red}{\textbf{86.80}}} & \rev{\textcolor{red}{\textbf{75.67}}} & \rev{\textcolor{red}{\textbf{85.62 $\pm$ 0.74}}} & \rev{\textcolor{red}{\textbf{66.68 $\pm$ 1.20}}} \\
    \bottomrule[2pt]
    \end{tabular}}%
  \label{tab:ID-results}%
\end{table*}%

\subsection{Implementation details}
Unless otherwise specified for baseline methods, we uniformly adopt the ViT-based pretrained model named BrainMVP~\cite{rui2024brainmvp} as the image feature encoder and the pretrained ClinicalBERT model~\cite{liu2025generalist} as the text encoder across all approaches, including our own.
\rev{BrainMVP is used through its ViT image-encoder branch: an input brain image is divided into patches, projected into token embeddings, and processed by stacked Transformer encoder blocks to obtain image-token features. We choose BrainMVP because it is pretrained on large-scale multi-parametric brain MRI data and is therefore well aligned with the neuroimaging inputs used for dementia etiology diagnosis. In our implementation, MRI and PET scans are passed through the same initialized encoder separately when present, providing a consistent image-feature space before COME performs modality-aware refinement and fusion. ClinicalBERT is a BERT-style medical language encoder used to encode standardized acquisition descriptions and categorical clinical variables. We choose it because these inputs contain medical terminology, abbreviations, and clinical semantics that are better represented by a domain-specific medical text encoder than by a general-domain language model. To ensure fair comparison, the same BrainMVP and ClinicalBERT initialization strategy is used for all compared methods that require image or text encoders, especially the multi-task learning baselines.}
The diagnosis model is trained for 100 epochs with a batch size of 8 using the Adam optimizer~\cite{kingma2014adam}. The initial learning rate is $1 \times 10^{-4}$, and is decayed by a factor of 0.1 at the 40th and 60th epochs. Additionally, a warm-up strategy is used with a learning rate of $1 \times 10^{-6}$ for the first 5 epochs to stabilize training.

\begin{table*}[htbp!]
  \centering
  \caption{Cross-Center and Cross-Sequence Generalization Performance. \rev{Cross-center results are reported as the mean over five random seeds, and cross-sequence results as mean $\pm$ standard deviation. Rec. denotes Recall-m.} Best results highlighted in bold red and second-best results underlined.}
    \setlength{\tabcolsep}{2pt}
    \renewcommand{\arraystretch}{1.15}
    \centering  \resizebox{\textwidth}{!}{
    \begin{tabular}{c|cccc|cccc|cccc|cccc|cccc}
    \toprule[2pt]
    \multirow{2}[1]{*}{Method} & \multicolumn{4}{c|}{ADNI} & \multicolumn{4}{c|}{NACC} & \multicolumn{4}{c|}{AIBL}  &\multicolumn{4}{c|}{OASIS} &\multicolumn{4}{c}{Cross-Sequence} \\
    \cmidrule{2-21}
    & AUC-m & mAP & F1-m & \rev{Rec.} & AUC-m & mAP & F1-m & \rev{Rec.} & AUC-m & mAP & F1-m & \rev{Rec.} & AUC-m & mAP & F1-m & \rev{Rec.} & AUC-m & mAP & F1-m & \rev{Rec.}\\
    \midrule[1.2pt]
    Radnet & \rev{\underline{79.62}} & \rev{\underline{62.69}} & \rev{52.70} & \rev{\underline{57.43}} & \rev{\underline{80.54}} & \rev{30.77} & \rev{24.12} & \rev{23.06} & \rev{74.08} & \rev{54.76} & \rev{37.89} & \rev{36.29} & \rev{54.91} & \rev{22.14} & \rev{\underline{13.62}} & \rev{17.44} & \rev{\underline{88.40 $\pm$ 0.44}} & \rev{\underline{33.79 $\pm$ 0.36}} & \rev{27.06 $\pm$ 0.75} & \rev{25.98 $\pm$ 0.95} \\
    M4Survive & \rev{77.61} & \rev{59.29} & \rev{47.27} & \rev{44.36} & \rev{74.84} & \rev{29.25} & \rev{25.39} & \rev{\underline{25.43}} & \rev{80.00} & \rev{61.12} & \rev{\underline{45.86}} & \rev{\underline{39.97}} & \rev{55.74} & \rev{23.98} & \rev{11.21} & \rev{17.10} & \rev{87.36 $\pm$ 0.19} & \rev{32.32 $\pm$ 0.34} & \rev{26.52 $\pm$ 0.63} & \rev{25.21 $\pm$ 0.65} \\
    NCOMMS22 & \rev{77.71} & \rev{60.64} & \rev{\underline{56.00}} & \rev{55.45} & \rev{79.90} & \rev{\underline{31.47}} & \rev{\underline{25.79}} & \rev{25.32} & \rev{\underline{80.50}} & \rev{\underline{61.14}} & \rev{45.63} & \rev{39.75} & \rev{\underline{57.74}} & \rev{23.66} & \rev{13.54} & \rev{\underline{18.54}} & \rev{86.63 $\pm$ 1.33} & \rev{32.46 $\pm$ 1.06} & \rev{26.99 $\pm$ 0.94} & \rev{\underline{26.34 $\pm$ 1.14}} \\
    NMED24 & \rev{79.48} & \rev{61.94} & \rev{49.40} & \rev{46.33} & \rev{68.67} & \rev{27.61} & \rev{22.92} & \rev{22.01} & \rev{62.92} & \rev{47.67} & \rev{37.48} & \rev{36.14} & \rev{55.74} & \rev{\textcolor{red}{\textbf{24.38}}} & \rev{11.21} & \rev{17.10} & \rev{86.66 $\pm$ 1.15} & \rev{33.43 $\pm$ 1.02} & \rev{\underline{27.11 $\pm$ 0.16}} & \rev{25.76 $\pm$ 0.53} \\
    \midrule[1.2pt]
    \rowcolor[rgb]{ .949,  .949,  .949} Ours & \rev{\textcolor{red}{\textbf{83.87}}} & \rev{\textcolor{red}{\textbf{67.29}}} & \rev{\textcolor{red}{\textbf{58.23}}} & \rev{\textcolor{red}{\textbf{58.35}}} & \rev{\textcolor{red}{\textbf{81.73}}} & \rev{\textcolor{red}{\textbf{32.46}}} & \rev{\textcolor{red}{\textbf{27.56}}} & \rev{\textcolor{red}{\textbf{28.14}}} & \rev{\textcolor{red}{\textbf{83.06}}} & \rev{\textcolor{red}{\textbf{65.73}}} & \rev{\textcolor{red}{\textbf{48.75}}} & \rev{\textcolor{red}{\textbf{45.01}}} & \rev{\textcolor{red}{\textbf{58.36}}} & \rev{\underline{24.22}} & \rev{\textcolor{red}{\textbf{17.12}}} & \rev{\textcolor{red}{\textbf{20.72}}} & \rev{\textcolor{red}{\textbf{89.46 $\pm$ 0.64}}} & \rev{\textcolor{red}{\textbf{36.13 $\pm$ 2.49}}} & \rev{\textcolor{red}{\textbf{27.81 $\pm$ 0.66}}} & \rev{\textcolor{red}{\textbf{27.27 $\pm$ 1.44}}} \\
    \bottomrule[2pt]
    \end{tabular}%
    }
  \label{tab:OOD-results}%
  \vspace{-12pt}
\end{table*}%

\subsection{In-Domain Performance}

To evaluate the robustness of COME to data heterogeneity across multiple centers, we conducted comprehensive in-domain experiments using seven independent datasets. Each dataset was split into training and testing subsets in a 4:1 ratio, with no patient overlap between splits. All models were trained on the aggregated multi-center data and evaluated separately on each center’s test set to measure domain-specific performance under heterogeneity conditions.

As shown in Table~\ref{tab:ID-results}, COME achieves optimal performance across most test datasets, demonstrating adaptation to diverse data characteristics, including varying population structures, acquisition protocols, and label distributions. \rev{In the AVG comparison, COME ranks first with 85.62 $\pm$ 0.74 AUC-m and 66.68 $\pm$ 1.20 mAP, outperforming the strongest baselines by +4.29\% AUC-m and +4.37\% mAP with statistically significant improvements on both metrics ($p<0.05$, paired two-sided t-test). At the dataset level, COME is particularly effective in challenging long-tailed settings; on PPMI, it outperforms Aligned by +6.69\% AUC-m and +12.64\% mAP.}
Notably, COME underperforms \rev{NMED24} on AIBL (\rev{89.00 vs. 90.75 AUC-m}). This is because COME is specifically designed to enhance robustness in multi-center, multi-protocol, and missing modalities scenarios, whereas AIBL constitutes a low-heterogeneity cohort (standardized protocols, complete modalities). This mismatch limits COME’s advantage and potentially requires further downstream adaptation.
Overall, these results substantiate that explicitly encoding meta knowledge through heterogeneity-aware embeddings enables the model to learn more transferable representations. The consistent performance gains across diverse evaluation contexts highlight our method's capacity to mitigate center shift.

\subsection{Out-of-domain performance}

To evaluate the generalization of our proposed method to unseen data distributions and acquisition protocols, we conducted two complementary out-of-domain (OOD) evaluations: cross-center validation and cross-sequence validation. 
The former assesses performance to unseen institutions, while the latter evaluates robustness to varying acquisition protocols.
\rev{For threshold-dependent metrics, including F1-m and Recall-m, we used a unified fixed threshold of 0.5 for all labels and methods.}
\rev{In practical deployment, this operating threshold can be further adjusted for each etiology according to the desired sensitivity-specificity trade-off.}

% These experiments evaluate the model’s ability to generalize in the presence of institutional and acquisition-related heterogeneity, which are critical challenges in real-world clinical deployment.

%只写两段，Cross-center Evaluation一段，包含实验设置和结果分析，Cross-sequence Evaluation类似，如果结论差不多，后一个的结果分析可以简写。
\subsubsection{Cross-center Evaluation}
We adopted a leave-one-center-out strategy using four large-scale datasets (ADNI, NACC, AIBL, OASIS) with sufficient sample diversity. For each test center, models were exclusively trained on data from the remaining centers. 
This setup simulates deployment to unseen institutions with varying demographics and protocols. 
As shown in Table~\ref{tab:OOD-results},  \rev{with cross-center results averaged over five random seeds,} our method consistently achieves state-of-the-art performance across all centers, demonstrating strong robustness and effective mitigation of center-specific biases through meta-knowledge enhancement.
\rev{In a seed-paired comparison with NCOMMS22, the strongest OOD baseline by mean rank, COME is higher in all 16 center-metric pairs across the four held-out centers, with 13 pairs significant under one-sided paired t-tests ($p<0.05$).}
\subsubsection{Cross-sequence Evaluation}
To specifically quantify resilience to MRI sequence heterogeneity, we isolated all samples acquired with two distinct sequences: "Sagittal\_3D\_FLAIR\_\_MSV21\_" and "MT1\_\_GradWarp\_\_N3m" (1,817 samples) as the test set. The training set comprised all other sequences (7,817 samples), creating a clinically relevant scenario where models must generalize across fundamentally different imaging characteristics without sequence-specific fine-tuning. As shown in Table~\ref{tab:OOD-results}, our model outperforms all baselines \rev{across four metrics} (+\rev{1.06}\% AUC-m, +\rev{2.34}\% mAP, \rev{+0.70\% F1-m, +0.93\% Recall-m}), highlighting the framework's ability to disentangle sequence-related heterogeneity from diagnostically relevant features, crucial for multi-center studies with non-standardized protocols. \rev{Against NCOMMS22, the strongest OOD baseline by mean rank, COME is higher in all four cross-sequence metrics, and all four improvements are significant under seed-paired one-sided t-tests ($p<0.05$).}
%

% \subsubsection{Results and Analysis:}

\begin{table*}[htbp!]
  \centering  
\caption{Improvement of meta knowledge enhancement. No Enhancement: meta information not used during inference. Improvement: Performance gain achieved with the enhancement. \rev{Asterisks indicate significant improvements ($p<0.05$); per-cohort columns use patient-level paired permutation tests, and AVG columns use dataset-level one-sided paired t-tests.}}
  \resizebox{\textwidth}{!}{
    \begin{tabular}{c|cc|cc|cc|cc|cc|cc|cc|cc}
    \toprule[2pt]
    \multirow{2}[2]{*}{Method} & \multicolumn{2}{c|}{PPMI} & \multicolumn{2}{c|}{NIFD} & \multicolumn{2}{c|}{4RTNI} & \multicolumn{2}{c|}{OASIS} & \multicolumn{2}{c|}{AIBL} & \multicolumn{2}{c|}{NACC} & \multicolumn{2}{c|}{ADNI} & \multicolumn{2}{c}{AVG} \\
        \cmidrule{2-17}
          & AUC-m   & mAP   & AUC-m   & mAP   & AUC-m   & mAP   & AUC-m   & mAP   & AUC-m   & mAP   & AUC-m   & mAP   & AUC-m   & mAP   & AUC-m   & mAP \\
    \midrule[1.2pt]
    No Enhancement & 84.70  & 72.06  & 96.57  & 96.61  & 68.87  & 62.12  & 65.97  & 31.08  & 87.42  & 72.75  & 87.93  & 41.48  & 86.17  & 75.23  & 82.52  & 64.48  \\
    Ours  & 95.71  & 81.35  & 97.30  & 97.45  & 76.57  & 69.80  & 71.00  & 33.96  & 88.50  & 74.06  & 89.83  & 42.93  & 86.84  & 76.20  & 86.54  & 67.96  \\
    \midrule[1.2pt]
    Improvement  & $\uparrow$ \textcolor{red}{\textbf{11.01\rev{\textsuperscript{*}}  }} & $\uparrow$ \textcolor{red}{\textbf{9.29\rev{\textsuperscript{*}}  }} & $\uparrow$ \textcolor{red}{\textbf{0.73  }} & $\uparrow$ \textcolor{red}{\textbf{0.84\rev{\textsuperscript{*}}  }} & $\uparrow$ \textcolor{red}{\textbf{7.69\rev{\textsuperscript{*}}  }} & $\uparrow$ \textcolor{red}{\textbf{7.68\rev{\textsuperscript{*}}  }} & $\uparrow$ \textcolor{red}{\textbf{5.03\rev{\textsuperscript{*}}  }} & $\uparrow$ \textcolor{red}{\textbf{2.88\rev{\textsuperscript{*}}  }} & $\uparrow$ \textcolor{red}{\textbf{1.08\rev{\textsuperscript{*}}  }} & $\uparrow$ \textcolor{red}{\textbf{1.31\rev{\textsuperscript{*}}  }} & $\uparrow$ \textcolor{red}{\textbf{ 1.90\rev{\textsuperscript{*}}  }} & $\uparrow$ \textcolor{red}{\textbf{ 1.45\rev{\textsuperscript{*}}  }} & $\uparrow$ \textcolor{red}{\textbf{ 0.67\rev{\textsuperscript{*}}  }} & $\uparrow$ \textcolor{red}{\textbf{ 0.97\rev{\textsuperscript{*}}  }} & $\uparrow$ \textcolor{red}{\textbf{4.02\rev{\textsuperscript{*}}  }} & $\uparrow$ \textcolor{red}{\textbf{ 3.49\rev{\textsuperscript{*}}}}  \\
    \bottomrule[2pt]
    \end{tabular}%
    }
  \label{tab:abla1}%
    \vspace{-6pt}
\end{table*}%

\begin{table}[tbp]
  \centering
  \caption{Ablation on Trust-Region Constraints}
  % \vspace{-3pt}
    \resizebox{0.33\textwidth}{!}{
    \begin{tabular}{cc|cc|cc}
    \toprule [2pt]
    \multirow{2}[4]{*}{$\mathcal{L}_g$} & \multirow{2}[4]{*}{$\mathcal{L}_l$} & \multicolumn{2}{c|}{ADNI} & \multicolumn{2}{c}{NACC} \\
\cmidrule{3-6}          &       & AUC-m & mAP   & AUC-m & mAP \\
    \midrule[1.2pt]
          $\times$ & $\times$      &  80.17     & 61.24      &  77.12     & 29.45 \\
          $\checkmark$ & $\times$  &  81.04     &  64.95     &  79.34     & 31.20 \\
          $\times$ & $\checkmark$      &  80.53     &   63.43    &  82.02     & 33.54 \\
    \midrule[1.2pt]
          $\checkmark$ & $\checkmark$       & 83.07 & 68.93 & 82.19 & 33.59 \\
    \bottomrule[2.0pt]
    \end{tabular}%
    }
  \label{tab:abla2}%
  \vspace{-12pt}
\end{table}%

\subsection{Ablation study}
\rev{We conduct ablation studies to assess the contribution of COME's key components and design choices, including meta-knowledge injection, constrained optimization, clinical/meta-information integration, and refinement/optimization settings. All ablations are reported under the same single-seed protocol to ensure fair comparisons.}

\subsubsection{Ablation on Meta information enhancement}
%
% To validate the adaptive capabilities of our knowledge enhancement mechanism, 
To validate the effectiveness of meta knowledge injection, we conducted an ablation experiment comparing model performance with and without meta knowledge enhancement during inference.
As evidenced by Table~\ref{tab:abla1}, the injection of information from acquisition protocols, cohorts, and modality-coverage substantially enhances diagnostic robustness across heterogeneous datasets.
\rev{At the AVG level, meta-knowledge enhancement improves AUC-m by +4.02\% and mAP by +3.49\%, with significant gains under dataset-level one-sided paired t-tests (AUC-m: $p=0.0197$; mAP: $p=0.0195$). Patient-level paired permutation tests further confirm significant improvements for all per-cohort entries except NIFD AUC-m.}
Our knowledge enhancement mechanism delivers dramatic improvements (+11.01\% AUC-m, +9.29\% mAP) on the extreme class-imbalanced dataset PPMI, confirming its ability to dynamically recalibrate decision boundaries for underrepresented etiologies. 
Similarly, substantial gains on small-scale dataset 4RTNI (+7.69\% AUC-m) and low-baseline OASIS (+5.03\% AUC-m) demonstrate the capacity of our method to amplify diagnostic signals in data-scarce environments through feature enhancement. 
%
% The variance in improvement magnitudes further reflects real-world diagnostic challenges: our method provides the greatest lift where conventional AI fails most severely (long-tailed distributions, sparse data).
%
Besides, the consistent but modest improvements in well-represented cohorts (e.g., +0.73\% AUC-m on NIFD) suggest our method operates as a precision refinement tool rather than merely compensating for deficiencies. 

\subsubsection{Ablation on Trust-Region Constraints}
%加入和没有constraint的对比，说明Trust-Region Constraints的提升性
Ablation studies in Table~\ref{tab:abla2} evaluate dual-level constraints by removing $\mathcal{L}_g / \mathcal{L}_l$, using ADNI/NACC as test sets with other centers for training. 
The results show that the full constraint system achieves substantial gains over the baseline (meta knowledge integration without constraints), improving AUC-m by +2.90\% on ADNI and +5.07\% on NACC. Moreover, despite the heterogeneous modality missing patterns in ADNI and high acquisition protocol variability in NACC (86 unique sequence names), the combined use of $\mathcal{L}_g$ and $\mathcal{L}_l$ jointly delivers consistent gains, demonstrating their complementary roles in reliable knowledge integration.
\rev{Paired permutation tests on matched test samples confirmed significant improvements in all ADNI comparisons and in the NACC comparisons against the no-constraint and global-only settings ($p<0.05$). The NACC comparison against the local-only variant was not significant, but the full model still achieved positive gains, suggesting that both constraints remain beneficial while the added global constraint has a smaller marginal effect in this split.}
Overall, the Trust-Region Constraints ensure the robustness and efficiency of meta knowledge injection.
\rev{We also evaluated the sensitivity of the two trust-region hyperparameters in the in-domain setting. With the default setting ($\lambda=0.2$, $k=2$), COME achieved 86.54 AUC-m and 67.96 mAP. Neighboring constraint weights $\lambda\in\{0.1,0.3\}$ gave slightly lower but comparable performance (85.35--85.42 AUC-m and 66.87--67.92 mAP), whereas a much stronger constraint ($\lambda=1$) reduced performance to 82.74 AUC-m and 65.48 mAP. For the reference update interval, $k=1$ gave 86.30 AUC-m/67.77 mAP and $k=3$ gave 86.49 AUC-m/68.02 mAP, indicating that COME is not sensitive to moderate changes in the hard-update frequency.}

\subsubsection{\rev{Contribution of clinical information}}
\rev{
To assess the role of clinical information, we compared COME with several variants under the same in-domain ablation setting in Table~\ref{tab:clinical_input_abla}. We first evaluated two clinical-only settings to separate clinical representation from imaging information: ``Raw only'' directly models the harmonized clinical values, whereas ``Emb. only'' uses the proposed clinical embedding branch without any image input. Therefore, Emb. only also serves as a clinical-information-only ablation. On AVG, Emb. only improves over Raw only from 77.66\% to 80.50\% AUC-m and from 58.25\% to 59.26\% mAP, suggesting that the clinical branch alone captures useful diagnostic signal and that embedding-based clinical modeling is more effective than directly using raw clinical values. We then evaluated whether raw or directly concatenated auxiliary information is sufficient in the multimodal setting. ``Image + raw'' combines imaging with raw clinical values, while ``Image + direct'' directly concatenates imaging with clinical and meta information without COME's structured embedding/refinement design. COME outperforms Image + raw by +5.04\% AUC-m and +6.73\% mAP on AVG, and also exceeds Image + direct by +3.33\% AUC-m and +3.35\% mAP. These results indicate that COME's improvement is not only due to the availability of clinical/meta information, but also to the structured embedding pathways and attention-based integration used to incorporate it.
}

\begin{table}[tbp]
  \centering
  \caption{\rev{Ablation on clinical information.}}
  \rev{
  \resizebox{0.45\textwidth}{!}{
    \begin{tabular}{l|cc|cc|cc}
    \toprule[2pt]
    \multirow{2}{*}{Method} & \multicolumn{2}{c|}{NACC} & \multicolumn{2}{c|}{ADNI} & \multicolumn{2}{c}{AVG} \\
    \cmidrule{2-7}
     & AUC-m & mAP & AUC-m & mAP & AUC-m & mAP \\
    \midrule[1.2pt]
    Raw only & 87.15 & 34.88 & 85.21 & 73.76 & 77.66 & 58.25 \\
    Emb. only & 88.72 & 35.86 & 85.63 & 74.85 & 80.50 & 59.26 \\
    \midrule[1.2pt]
    Image + raw & 87.90 & 39.10 & 84.72 & 74.72 & 81.50 & 61.23 \\
    Image + direct & 88.42 & 38.23 & 86.29 & 75.36 & 83.21 & 64.61 \\
    \midrule[1.2pt]
    \rowcolor[rgb]{ .949, .949, .949} COME & \textbf{89.83} & \textbf{42.93} & \textbf{86.84} & \textbf{76.20} & \textbf{86.54} & \textbf{67.96} \\
    \bottomrule[2pt]
    \end{tabular}}
  }
  \label{tab:clinical_input_abla}
  \vspace{-12pt}
\end{table}

\begin{table}[tbp]
  \centering
  \caption{\rev{Ablation on architectural and optimization choices in the in-domain setting. AVG denotes the average over the seven datasets.}}
  \rev{
  \resizebox{0.45\textwidth}{!}{
    \begin{tabular}{l|l|cc}
    \toprule[2pt]
    Block & Variant & AVG AUC-m & AVG mAP \\
    \midrule[1.2pt]
    \multirow{4}{*}{Depth}
    & $J=1$ & 83.27 & 65.43 \\
    & $J=3$ & \textbf{87.18} & 66.72 \\
    & $J=4$ & 86.31 & 67.28 \\
    & $J=2$ (default) & 86.54 & \textbf{67.96} \\
    \midrule[1.2pt]
    \multirow{3}{*}{Loss}
    & All contrastive & 85.49 & 64.86 \\
    & All cosine-MSE & 86.07 & 65.91 \\
    & Hybrid (default) & \textbf{86.54} & \textbf{67.96} \\
    \midrule[1.2pt]
    \multirow{3}{*}{Schedule}
    & No init. & 81.63 & 60.58 \\
    & Loss ramp & 85.22 & 65.74 \\
    & Two phase (default) & \textbf{86.54} & \textbf{67.96} \\
    \bottomrule[2pt]
    \end{tabular}}
  }
  \label{tab:arch_opt_choices}
  \vspace{-12pt}
\end{table}

\subsubsection{\rev{Architectural and optimization choices}}
\rev{
We further examined three architectural and optimization choices in the in-domain setting, as summarized in Table~\ref{tab:arch_opt_choices}.
First, increasing the cross-attention refinement depth from $J=1$ to the default $J=2$ improved AVG AUC-m from 83.27\% to 86.54\% and mAP from 65.43\% to 67.96\%, indicating that a single meta-knowledge refinement layer is insufficient. Further increasing the depth to $J=3$ or $J=4$ did not yield consistent gains across both metrics, suggesting that two refinement layers provide a suitable balance between progressive modulation and model complexity.
Second, the hybrid alignment design, which uses minibatch contrastive global feature alignment and cosine-affinity MSE for local patch alignment, outperformed using the same contrastive or cosine-MSE formulation for both levels. This supports the use of contrastive patient-level alignment globally and structure-preserving affinity alignment locally. 
Third, the two-phase training strategy outperformed two one-phase alternatives. ``No init.'' removes the initialization phase and trains the reference model from random parameters, while ``Loss ramp'' gradually increases the trust-region loss weight from 0 to 0.2 over the first 20 epochs. Both variants underperformed the default setting, supporting first learning a diagnosis-oriented representation from core clinical inputs and then applying constrained meta-knowledge enhancement.
}

\subsubsection{\rev{Representation analysis of meta-knowledge embeddings}}
\begin{figure}[t!]
\centerline{\includegraphics[width=0.98\linewidth]{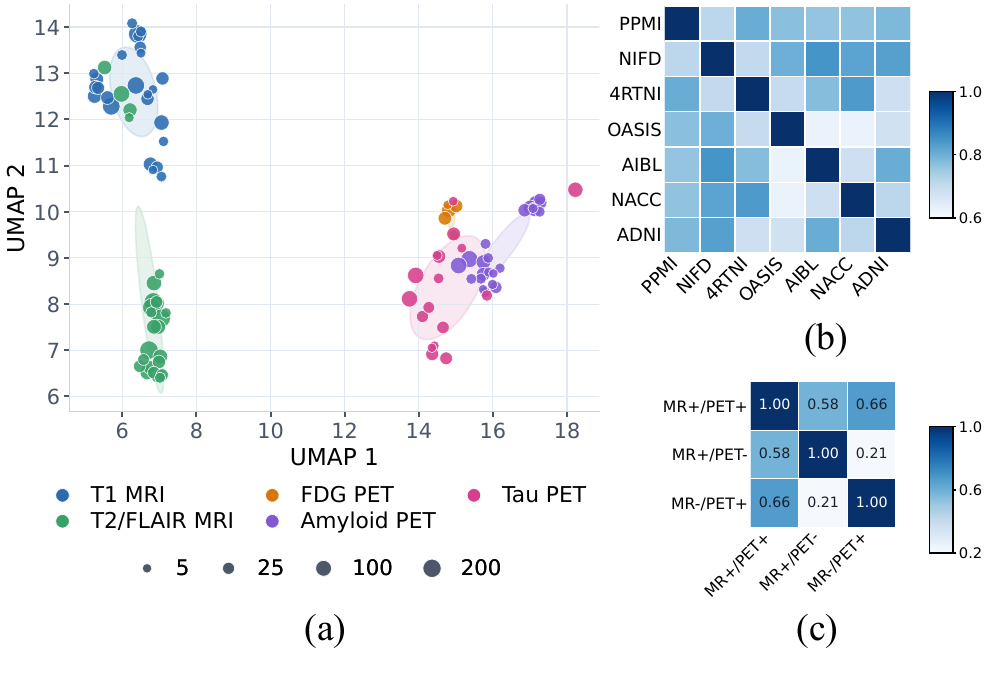}}
\vspace{-6pt}
\caption{\rev{Representation analysis of meta-knowledge embeddings in NC samples. (a) UMAP of acquisition-level sequence representations. (b) Cohort-token centroid cosine similarity. (c) Modality-coverage token centroid cosine similarity.}}
\label{fig:meta_embedding_analysis}
\vspace{-6pt}
\end{figure}

\rev{To examine whether the learned meta-knowledge tokens encode the intended heterogeneity information after training, we analyzed the trained token representations used by COME. To reduce confounding from diagnostic-label differences, this analysis used NC samples and compared variation across acquisition names, cohort sources, and modality-coverage patterns. As shown in Fig.~\ref{fig:meta_embedding_analysis}~(a), acquisition-level embeddings were obtained by averaging L2-normalized sequence representations for each acquisition name and visualized with UMAP. The points form distinguishable coarse semantic groups, including T1 MRI, T2/FLAIR MRI, FDG PET, amyloid PET, and tau PET. This suggests that the acquisition-aware tokens preserve major protocol-level semantics rather than behaving as arbitrary sequence identifiers. Fig.~\ref{fig:meta_embedding_analysis}~(b)--(c) further shows centroid cosine-similarity heatmaps for sample-conditioned cohort injected tokens and modality-coverage tokens. Cohort tokens show source-dependent similarity structure, while modality-coverage tokens clearly separate MR/PET present-missing patterns. Together, these results support that the trained meta-knowledge embeddings encode the intended acquisition, cohort, and modality-availability information used for attention-based feature refinement.}

\begin{figure}[ht!]
\centerline{\includegraphics[width=\linewidth]{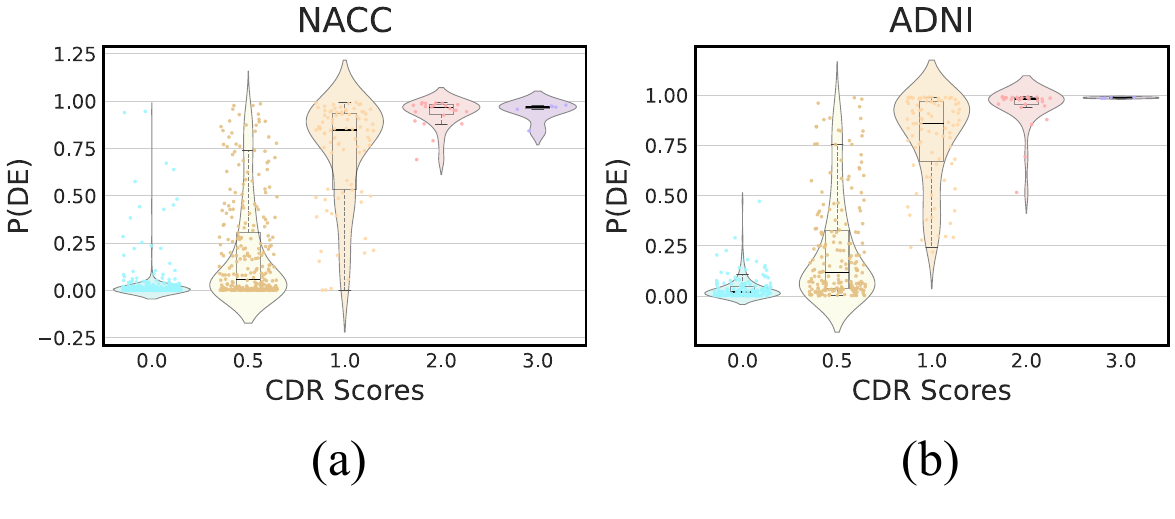}}

\caption {Clinical indicator level verification. Raincloud plots are shown to denote the distribution of CDR scores (x axis) versus model-predicted probability of dementia (y axis), on the NACC and ADNI cohorts, respectively. We performed the Kruskal-Wallis H-test for independent samples in NACC (n = 1017, H = 453.90, P = 6.25e-97), ADNI (n = 643, H = 301.36, P = 5.51e-64).}

\label{fig:cdr}
\vspace{-12pt}
\end{figure}

\begin{figure*}[t!]
\centerline{\includegraphics[width=\linewidth]{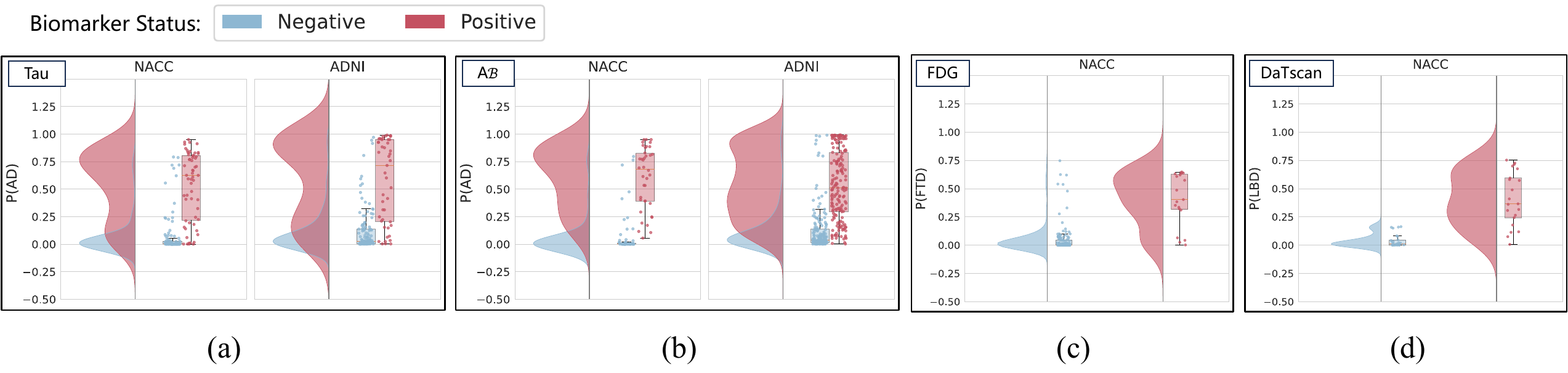}}
\vspace{-7pt}
\caption{Biomarker level validation. Raincloud plots representing model probabilities for dementia etiologies across their respective biomarker-negative (blue) and positive groups (red). (a): P(AD) in relation to tau PET; (b): P(AD) in relation to amyloid \(\beta\) (A\(\beta\)); (c): P(FTD) in relation to FDG PET; (d): P(LBD) in relation to DaTscan.}
\label{fig:biomarker}
\vspace{-7pt}
\end{figure*}

\begin{figure*}[ht!]
\centerline{\includegraphics[width=0.98\linewidth]{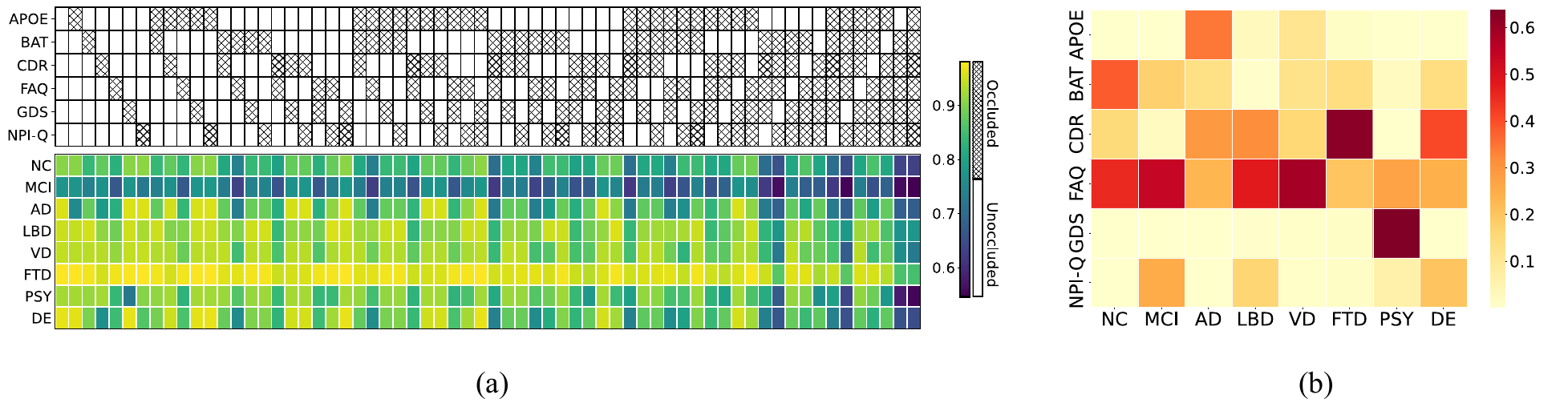}}
\vspace{-8pt}
\caption {Feature occlusion analysis and derived feature importance. (a) Feature occlusion patterns and corresponding model performance. Top: Visualization of occlusion configurations across all input features; hatched squares denote occluded features. Bottom: Corresponding model performance (AUC) under each configuration, with color intensity indicating performance. (b) Normalized feature importance heatmap. Relative importance of six clinical features (y-axis) for predicting eight dementia-related diagnostic categories (x-axis), derived from performance drops in (a). Importance scores represent the normalized average AUC decrease upon occlusion (darker = higher importance), scaled to sum to 1 within each diagnostic category. Analysis based on the top 1,000 cases with the most complete feature sets.}
\label{fig:Importance+drop}
\vspace{-14pt}
\end{figure*}

\subsection{Multi-level validation}
To comprehensively validate our model’s clinical and pathological relevance, we conducted two-tiered correlation analyses:
\subsubsection{Clinical-level validation} 
We evaluated the alignment between our model's predicted dementia probability (P(DE), generated without CDR inputs) and Clinical Dementia Rating (CDR) scores. Raincloud plots (Fig.~\ref{fig:cdr}) revealed significant positive correlations in both the NACC (n = 1017, H = 453.90, P = \rev{6.25e-97}) and ADNI (n = 643, H = 301.36, P = 5.51e-64) cohorts, confirming that our predictions capture established gradations in clinical dementia severity. Analysis showed P(DE) increased progressively with higher CDR scores in NACC and ADNI, P(DE) exhibited significant demarcation between baseline CDR and higher severity levels, highlighting the model's sensitivity to incremental impairment. Collectively, these results demonstrate the model's robust capacity to differentiate cognitive states.

\subsubsection{Biomarker-level validation}
We assessed the correspondence between model-predicted probabilities for P(AD) (alzheimer’s disease), P(FTD) (frontotemporal dementia), P(LBD) (Lewy body dementia), and the presence of associated biomarkers, as illustrated in the raincloud plots in Fig.~\ref{fig:biomarker}.
In both the NACC and ADNI cohorts, P(AD) showed significant associations with amyloid \(\beta\) (A\(\beta\)) and tau PET biomarkers, with highly significant differences observed between biomarker-positive and biomarker-negative groups ($P < 0.0001$). Notably, P(AD) was consistently elevated in individuals who tested positive for A\(\beta\) and tau, indicating that our diagnostic framework is well-aligned with the established amyloid and tau criteria for Alzheimer’s disease diagnosis 
~\cite{han2012beta}. Similarly, in the NACC cohort, P(FTD) and P(LBD) also demonstrated strong alignment with their respective biomarkers (FDG PET and DaTscan), further validating the biological plausibility of our model predictions.

%这里可以加一个引用 

%这个小节太长了，建议分成两个子节，每个子节两段，第一段写实验设置，第二段分析。另外这个实验我没咋看出他对你方法有效性的证明作用，如果你是想说明他符合诊断逻辑，建议的写法是，xxx疾病在诊断中关注xxx性质【引用】，图中xxx结果正好与之相符。
\subsection{Feature Importance Analysis via Occlusion}
To quantitatively evaluate the influence of key clinical and cognitive assessment metrics on the model's diagnostic decisions across different etiologies, we conducted a systematic occlusion-based feature importance analysis.

% \subsubsection{Occlusion-based Feature Importance Visualisation}
We specifically focused on six critical input indicators derived from common assessments: APOE ($\epsilon$4 genotype), BAT (Brief Cognitive Test), CDR (Clinical Dementia Rating), FAQ (Functional Activities Questionnaire), GDS (Geriatric Depression Scale), and NPI-Q (Neuropsychiatric Inventory Questionnaire).
%
% To minimize the impact of missing features in the dataset, we first ranked all cases based on the completeness of their available features. 
% We then selected the top 1000 cases with the most complete feature sets for further analysis. 
To ensure data reliability, we selected the top 1000 cases with the most complete feature profiles.
Next, we systematically ablated these six features across all possible combinations, resulting in $2^6 = 64$ distinct input configurations per diagnostic sample (each feature being either present or occluded).
Model performance (AUC) was recorded for all eight diagnostic categories (NC, MCI, AD, LBD, VD, FTD, PSY, DE) under each configuration (Fig.~\ref{fig:Importance+drop} (a)). 

Feature importance score was quantified as the average drop in performance when a specific feature was occluded, with scores normalized per category for cross-disease comparison. These normalized importance scores are presented as a heatmap (Fig.~\ref{fig:Importance+drop} (b)). Analysis of the heatmap reveals etiology-specific feature dependencies consistent with established clinical knowledge.
For Alzheimer’s disease (AD), APOE $\epsilon$4 genotype showed the strongest positive influence, aligning with its status as the strongest genetic risk factor according to the National Institute on Aging–Alzheimer’s Association (NIA-AA) guidelines~\cite{jack2018nia}.
For psychiatric disease-associated dementia (PSY), particularly depression and schizophrenia, the GDS emerged as the most influential factor, consistent with depression screening in dementia differentials~\cite{goodarzi2017depression}.
These results validate that our model intrinsically prioritizes features with established clinical relevance, indicating that the model’s data-driven diagnostic reasoning aligns with medical guidelines without relying on prior knowledge.

%% file: body/conclusion.tex
\section{conclusion}
In this work, we introduced the Collaborative Meta Knowledge Enhancement (COME) framework to address the combined challenges of acquisition heterogeneity, cohort
heterogeneity, and modality-coverage heterogeneity in multi-center dementia diagnosis. 
By explicitly encoding acquisition semantics, data source, and modality indicators as diverse embeddings, and regularizing their integration via trust-region constraints, our method achieves robust representation learning while mitigating spurious correlations.
Extensive evaluations across seven independent cohorts demonstrated that our approach consistently outperforms state-of-the-art baselines in both in-domain and out-of-domain settings.
Multi-level validations further confirmed the clinical validity of our predictions through strong correlations with established biomarkers and clinical indicators.
These results suggest that our framework can serve as a powerful and interpretable strategy for real-world, multi-center dementia diagnostics.
\rev{Despite these promising results, COME remains limited by the quality and completeness of meta-knowledge, the underrepresentation of rare dementia etiologies in retrospective datasets, and the lack of prospective clinical validation. Future work will improve metadata quality control, expand balanced multi-center cohorts, and conduct prospective validation in collaboration with neurology departments.}

% \rev{Despite these promising results, several limitations remain. First, COME relies on the availability and quality of meta knowledge, including acquisition descriptions and modality-coverage indicators; inaccurate sequence standardization or incomplete metadata may weaken the benefit of knowledge enhancement. Second, rare dementia subtypes such as LBD, VD, FTD, and PSY are underrepresented in the public retrospective datasets used here, calling for larger and more balanced multi-center validation. Finally, while biomarker- and clinical-level validations support the plausibility of the model predictions, they do not replace prospective clinical validation.}
% %
% % Future work will engage with neurology departments in hospitals to conduct comprehensive clinical validations of our model.
% \rev{Future work will focus on improving metadata quality control, expanding balanced cohorts for rare dementia etiologies, and conducting comprehensive clinical validations in collaboration with neurology departments.
% }

%% file: body/appendix.tex
\appendices
\section{\rev{LLM-Based Acquisition-Sequence Standardization}}
\label{app:seq_standardization}

\rev{This appendix provides supplementary details on acquisition-sequence standardization. The purpose of this step is to convert heterogeneous raw sequence names into concise technical descriptions with a fixed prompt, so that different cohort-specific naming conventions can be represented in a consistent textual form. Table~\ref{tab:appendix_seq_examples} shows representative examples of raw sequence names and their standardized descriptions.}

\rev{The prompt template used for DeepSeek~V3 was:}
\begin{quote}
\footnotesize
\ttfamily
\rev{For the given brain medical imaging acquisition sequence, provide a concise professional description of its key imaging characteristics and typical clinical use.}\\
\rev{Requirements:}\\
\rev{- Use no more than 30 English words.}\\
\rev{- Do not include the sequence name itself.}\\
\rev{- Keep the description technically accurate and domain-appropriate.}\\
\rev{- Return only the description, with no extra explanation, labels, or punctuation-heavy formatting.}\\
\rev{Example output:}\\
\rev{T1-weighted anatomical imaging with high spatial resolution for structural assessment and cortical or subcortical morphometry.}\\
\rev{Sequence: \{sequence\_name\}}
\end{quote}

\begin{table}[tb]
\centering
\caption{\rev{Examples of LLM-based acquisition-sequence standardization.}}
\label{tab:appendix_seq_examples}
{\scriptsize
\setlength{\tabcolsep}{3pt}
\renewcommand{\arraystretch}{1.10}
\begin{tabularx}{\columnwidth}{@{}>{\raggedright\arraybackslash}p{0.30\columnwidth}>{\raggedright\arraybackslash}X@{}}
\toprule[1.5pt]
\rev{\textbf{Raw sequence name}} & \rev{\textbf{DeepSeek~V3 output}} \\
\midrule[1.2pt]
\rev{\texttt{MR\_Space\_3D\_T2}\newline
\texttt{\_FLAIR\_sag\_p2}} &
\rev{3D T2 FLAIR imaging with CSF suppression, enabling high-resolution sagittal brain assessment of white matter lesions, cortical abnormalities, edema, and inflammatory or demyelinating pathology.} \\
\rev{\texttt{Sagittal\_3D\_FLAIR}\newline
\texttt{\_\_MSV21\_}} &
\rev{Sagittal 3D fluid-attenuated imaging suppressing CSF signal, useful for detecting white matter lesions, cortical abnormalities, demyelination, edema, and structural pathology.} \\
\rev{\texttt{MT1\_\_GradWarp}\newline
\texttt{\_\_N3m}} &
\rev{T1-weighted structural MRI corrected for gradient distortion and intensity nonuniformity, used for anatomical assessment, brain morphometry, and cortical or subcortical volume analysis.} \\
\rev{\texttt{PET\_DYNAMIC\_PET}\newline
\texttt{\_Brain\_75MIN}} &
\rev{Dynamic PET acquisition for quantitative brain tracer kinetics, enabling time activity curve analysis, parametric imaging, and assessment of cerebral metabolism or receptor binding.} \\
\bottomrule[1.5pt]
\end{tabularx}
}
\end{table}

\begin{table}[h!]
\centering
\footnotesize
\caption{\rev{Dataset-specific diagnosis sources used for label harmonization.}}
\label{tab:appendix_label_harmonization}
\setlength{\tabcolsep}{3pt}
\renewcommand{\arraystretch}{1.15}
\begin{tabularx}{\columnwidth}{@{}>{\raggedright\arraybackslash}p{0.13\columnwidth}>{\raggedright\arraybackslash}X@{}}
\toprule[1.5pt]
\rev{\textbf{Dataset}} & \rev{\textbf{Diagnosis-source / mapping basis}} \\
\midrule[1.2pt]
\rev{NACC} &
\rev{UDS clinician diagnosis fields, supplemented by FTLD/LBD module information when available; source-documented etiologies were mapped to the harmonized dementia categories.} \\
\rev{ADNI} &
\rev{ADNI clinical diagnostic cohort assignments and clinical protocol fields; NC, MCI, and AD dementia labels were mapped, while non-AD dementia etiologies were not assigned without an official source diagnosis.} \\
\rev{AIBL} &
\rev{Source-cohort clinical diagnosis fields from AIBL; NC, MCI, and AD dementia labels were mapped according to the cohort-provided clinical classification.} \\
\rev{PPMI} &
\rev{PPMI protocol and clinical case-report fields for Parkinson's disease, control/prodromal status, and cognitive assessments; no dementia-etiology label was mapped without an official source etiology field.} \\
\rev{OASIS} &
\rev{Knight ADRC/OASIS clinical diagnosis fields and CDR-based clinical assessments; documented dementia diagnoses were mapped when they corresponded to the harmonized categories.} \\
\rev{NIFD} &
\rev{FTLDNI/NIFD clinical diagnoses for frontotemporal lobar degeneration spectrum disorders; eligible diagnoses were mapped to the FTD-related category.} \\
\rev{4RTNI} &
\rev{4RTNI clinical diagnoses for 4-repeat tauopathy spectrum disorders, including corticobasal and progressive supranuclear palsy phenotypes; eligible diagnoses were mapped to the FTD-related category.} \\
\bottomrule[1.5pt]
\end{tabularx}
\end{table}

\section{\rev{Dataset-Specific Label Harmonization}}
\label{app:label_harmonization}

\rev{Table~\ref{tab:appendix_label_harmonization} summarizes the dataset-specific diagnosis sources used to map cohort-provided clinical labels to the harmonized diagnostic schema defined in Table~\ref{tab:etiology_definitions}, based on the corresponding dataset descriptions~\cite{beekly2004national,mueller2005ways,ellis2006australian,marek2011parkinson,marcus2010open,boxer2013frontotemporal,dutt2016progression}. The table is intentionally concise, as label harmonization relied only on source-documented diagnosis fields rather than re-diagnosing participants from imaging findings or biomarker patterns.}

\vspace{1cm}

%% file: body/ref.tex
{
\bibliographystyle{IEEEtran}
% \bibliography{reference.bib}
\bibliography{IEEEabrv,reference} 
}